\documentclass[11pt,a4paper]{article}
\usepackage[hyperref]{acl2018}
\usepackage{times}
\usepackage{latexsym}
\usepackage{mathptmx}
\usepackage{url}

\usepackage{hyperref}
\hypersetup{hidelinks}
\usepackage[hyphenbreaks]{breakurl}
\usepackage{xcolor}
\usepackage{bibentry}
\usepackage{graphicx}
\usepackage{stfloats}
\usepackage{subfigure}
\usepackage{multirow}
\usepackage{amssymb}
\usepackage{amsmath}
\allowdisplaybreaks
\usepackage{array}
\usepackage{algorithm}
\usepackage{algorithmic}

\DeclareMathAlphabet{\mathcal}{OMS}{cmsy}{m}{n}
\newcommand{\tabincell}[2]{\begin{tabular}{@{}#1@{}}#2\end{tabular}}
\newcolumntype{L}[1]{>{\raggedright\arraybackslash}p{#1}}
\newcolumntype{C}[1]{>{\centering\arraybackslash}p{#1}}
\newcolumntype{R}[1]{>{\raggedleft\arraybackslash}p{#1}}

\aclfinalcopy 

\title{Improving Aspect Term Extraction with Bidirectional \\Dependency Tree Representation}

\author{Huaishao Luo$^1$, Tianrui Li$^{*1}$, Bing Liu$^2$, Bin Wang$^1$, \and Herwig Unger$^3$\\
	$^1$School of Information Science and Technology, Southwest Jiaotong University \\
	{\tt huaishaoluo@gmail.com, trli@swjtu.edu.cn, binwang007@gmail.com} \\
	$^2$Department of Computer Science, University of Illinois at Chicago \\
	{\tt liub@uic.edu} \\
	$^3$Faculty of Mathematics and Computer Science, Fern University in Hagen \\
	{\tt herwig.unger@gmail.com}
}

\date{}
\begin{document}
	
	\maketitle
	
	\begin{abstract}
		Aspect term extraction is one of the important subtasks in aspect-based sentiment analysis. Previous studies have shown that using dependency tree structure representation is promising for this task. However, most dependency tree structures involve only one directional propagation on the dependency tree. In this paper, we first propose a novel bidirectional dependency tree network to extract dependency structure features from the given sentences. The key idea is to explicitly incorporate both representations gained separately from the bottom-up and top-down propagation on the given dependency syntactic tree. An end-to-end framework is then developed to integrate the embedded representations and BiLSTM plus CRF to learn both tree-structured and sequential features to solve the aspect term extraction problem. Experimental results demonstrate that the proposed model outperforms state-of-the-art baseline models on four benchmark SemEval datasets.
	\end{abstract}
	
	\section{Introduction}
	\label{sec:introduction}
	Aspect term extraction (ATE) is the task of extracting the attributes (or aspects) of an entity upon which people have expressed opinions. It is one of the most important subtasks in aspect-based sentiment analysis \cite{Liu2012}. As examples shown in Table \ref{table-example}, ``design'', ``atmosphere'', ``staff'', ``bar'', ``drinks'', and ``menu'' in the first two sentences are aspect terms of the restaurant reviews, and ``operating system'', ``preloaded software'', ``hard disc'', ``windows'', and ``drivers'' in the last two sentences are aspects terms of the laptop reviews.
	
	Existing methods for ATE can be divided into unsupervised and supervised approaches. The unsupervised approach is mainly based on topic modeling \cite{Lin2009,Brody2010,Moghaddam2011,Chen2013,Chen2014Topic,Chen2014Aspect}, syntactic rules \cite{Wang2008,Zhang2010,Wu2009,Qiu2011,Liu2013}, and lifelong learning \cite{Chen2014Aspect,Wang2016,Liu2016a,Shu2017}. The supervised approach is mainly based on Conditional Random Fields (CRF) \cite{Lafferty2001,Jakob2010,Choi2010,Li2010a,Mitchell2013a,Giannakopoulos2017Unsupervised}.
	
	\begin{table}[tp]
		\caption{\label{table-example} Example of user' review with aspect term marked in bold.}
		\begin{center}
			\begin{tabular}{|C{0.4cm}|p{6.0cm}|}
				\hline
				\textbf{No.} & \multicolumn{1}{l|}{\textbf{Reviews}} \\ \hline \hline
				1 & The \textbf{design} and \textbf{atmosphere} are just as good. \\
				2 & The \textbf{staff} is very kind and well trained, they're fast, they are always prompt to jump behind the \textbf{bar} and fix \textbf{drinks}, they know details of every item in the \textbf{menu} and make excellent recommendation. \\ 
				3 & I love the \textbf{operating system} and the \textbf{preloaded software}. \\
				4 & There also seemed to be a problem with the \textbf{hard disc}, as certain times \textbf{windows} loads but claims to not be able to find any \textbf{drivers} or files. \\ \hline
			\end{tabular}
		\end{center}
	\end{table}
	This paper focuses on CRF-based models, which regard ATE as a sequence labeling task. There are three main types of features that have been used in previous CRF-based models for ATE. The first type is the traditional natural language features, e.g., syntactic structures and lexical features \cite{Toh2016,Hamdan2015,Toh2015,balage2014nilc_usp,Jakob2010,Shu2017}. The second type is the cross domain knowledge based features, which are useful because there are plenty of shared aspects across domains although each entity/product is different \cite{Jakob2010,Mitchell2013a,Shu2017}. The final type is the deep learning features learned by deep learning models, which have been proven very useful for the ATE in recent years \cite{Giannakopoulos2017Unsupervised,Liu2015FineGrained,Wang2016a,Yin2016,Ye2017,Li2017Deep,Wang2017,Wang2017a}.
	
	\begin{figure*}[t]
		\centering
		\includegraphics[width=.85\textwidth]{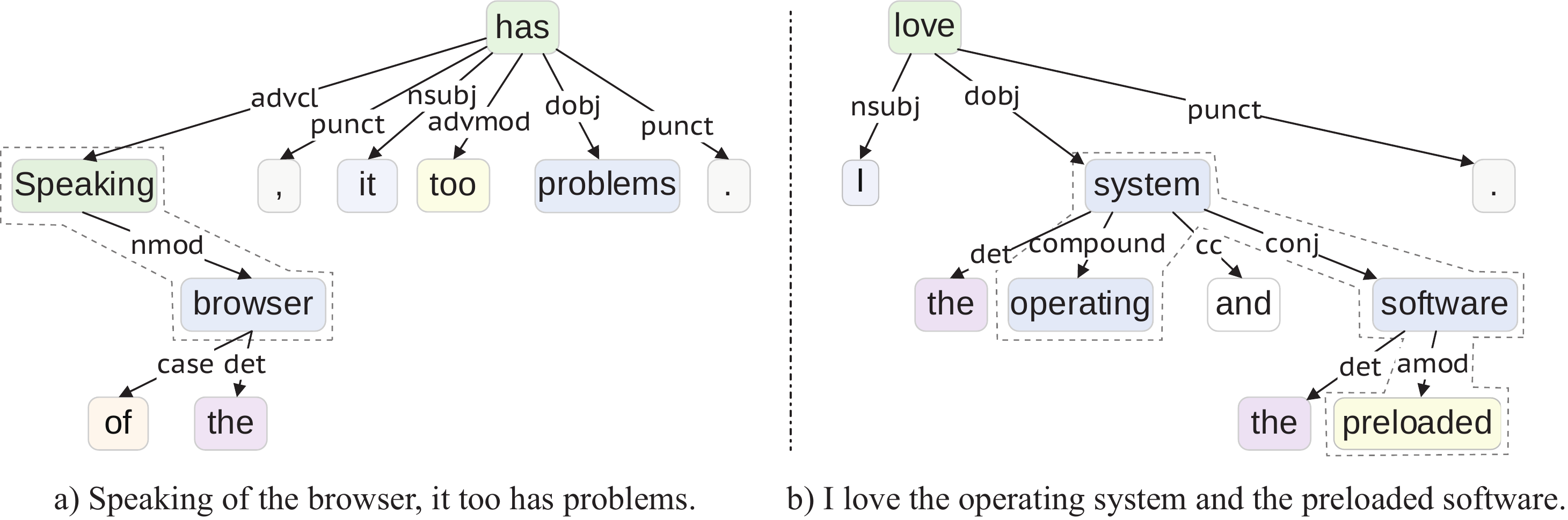}
		\caption{Examples of dependency relations (generated by the basic dependencies of Stanford CoreNLP 3.8.0). Each node is a word, and each edge is the dependency relation between two words.}\label{fig_dependency_tree}
	\end{figure*}
	The deep learning features generally include sequential representation and tree-structured representation features. Sequential representation means the word order of a sentence. Tree-structured representation features come from the syntax structure of a sentence, which represent the internal logical relations between words. Figure \ref{fig_dependency_tree} shows two examples of the dependency structure, in which each node is a word of the sentence, and each edge is a dependency relation between words. For example, the relation $Speaking\xrightarrow{nmod}browser$ means $Speaking$ is a nominal modifier of $browser$. Such a relation is useful in ATE. For instance, given \emph{system} as an aspect term, \emph{software} can be extracted as an aspect term through the relation: $system\xrightarrow{conj}software$ in Figure \ref{fig_dependency_tree} b) because $conj$ means $system$ and $software$ are connected by a coordinating conjunction (e.g., \emph{and}). However, the tree-structured representation in the previous work only considered a single direction of propagation (bottom-up propagation) trained on the parse trees with shared weights. We further exploit the capability of the tree-structured representation by considering top-down propagation, which means that given \emph{software} as an aspect term, \emph{system} can be extracted as an aspect term through the relation: $software\xrightarrow{conj^{-1}}system$, where $conj^{-1}$ is the inverse relation of the $conj$ for the purpose of distinguishing different directions of propagation. Compared with the sequential representation, the tree-structured representation is capable of obtaining the long-range dependency relation between words, especially for long sentences like the second and fourth reviews in Table \ref{table-example}.
	
	\begin{figure}[t]
		\centering
		\includegraphics[width=.40\textwidth]{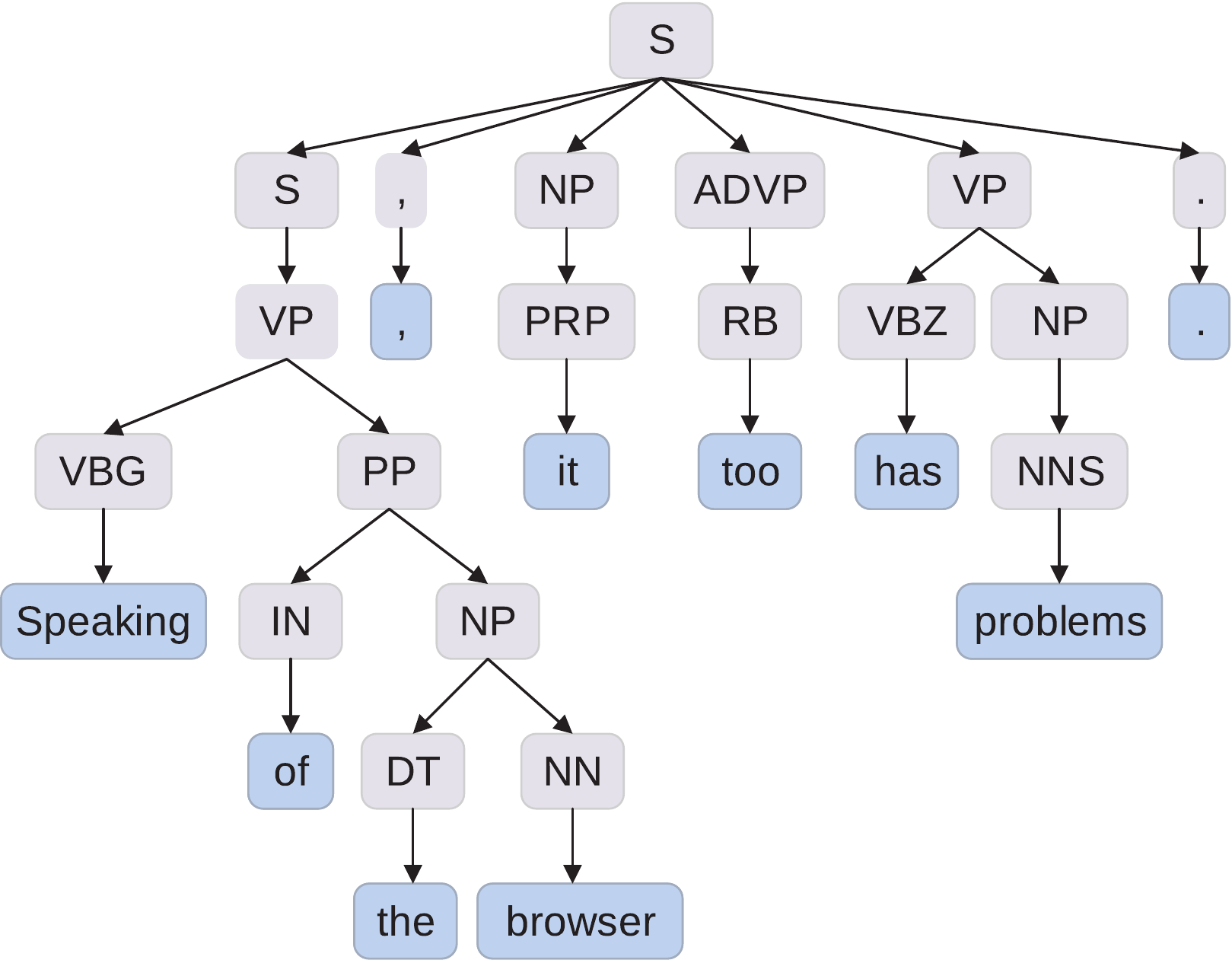}
		\caption{An example of a constituency tree (generated by the constituency parse of Stanford CoreNLP 3.8.0). Each node with the blue background is a real word in the sentence: \emph{Speaking of the browser, it too has problems.}}\label{fig_constituency_tree}
	\end{figure}
	In this paper, we first enhance the tree-structured representation using a bidirectional gate control mechanism which originates from bidirectional LSTM (BiLSTM) \cite{Hochreiter1997,Gers1999} and then fuse the tree-structured and the sequential information to perform the aspect term extraction. By combining the two steps into one, we propose a novel framework named \underline{bi}directional \underline{d}ependency \underline{tree} \underline{c}onditional \underline{r}andom \underline{f}ields (BiDTreeCRF). Specifically, BiDTreeCRF is an incremental framework, which consists of three main components. The first component is a \underline{bi}directional \underline{d}ependency \underline{tree} network (BiDTree), which is an extension of the recursive neural network in \cite{Socher2011}. Its goal is to extract the tree-structured representation from the dependency tree of a given sentence. The second component is the BiLSTM, whose input is the output of BiDTree. The tree-structured and sequential information is fused in this layer. The last component is the CRF, which is used to generate labels. To the best of our knowledge, this is the first work to fuse tree-structured and sequential information to solve the ATE. This new model results in major improvements for ATE over the existing baseline models.
	
	The proposed BiDTree is constructed based on the dependency tree. Compared with many other methods based on the constituency tree (Figure \ref{fig_constituency_tree}) \cite{Irsoy2013,Tai2015,Teng2016,Chen2017}, BiDTree focuses more directly on the dependency relation between words because all nodes in the dependency tree are input words themselves, but the constituency tree focuses on identified phrases and their recursive structure.
	
	The two main contributions of this paper are as follows. 
	
	\begin{itemize}
		\item It proposes a novel bidirectional recursive neural network BiDTree, which enhances the tree-structured representation by constructing a bidirectional propagation mechanism on the dependency tree. Thus, BiDTree can capture more effective tree-structured representation features and gain better performance.
		\item It proposes the incremental framework BiDTreeCRF, which can incorporate both the syntactic information and the sequential information. These pieces of information are fed into the CRF layer for aspect term extraction. The integrated model can be effectively trained in an end-to-end fashion.
	\end{itemize}
	
	\section{Model Description}
	\label{sec_model}
\begin{figure*}
		\centering
		\includegraphics[width=.96\textwidth]{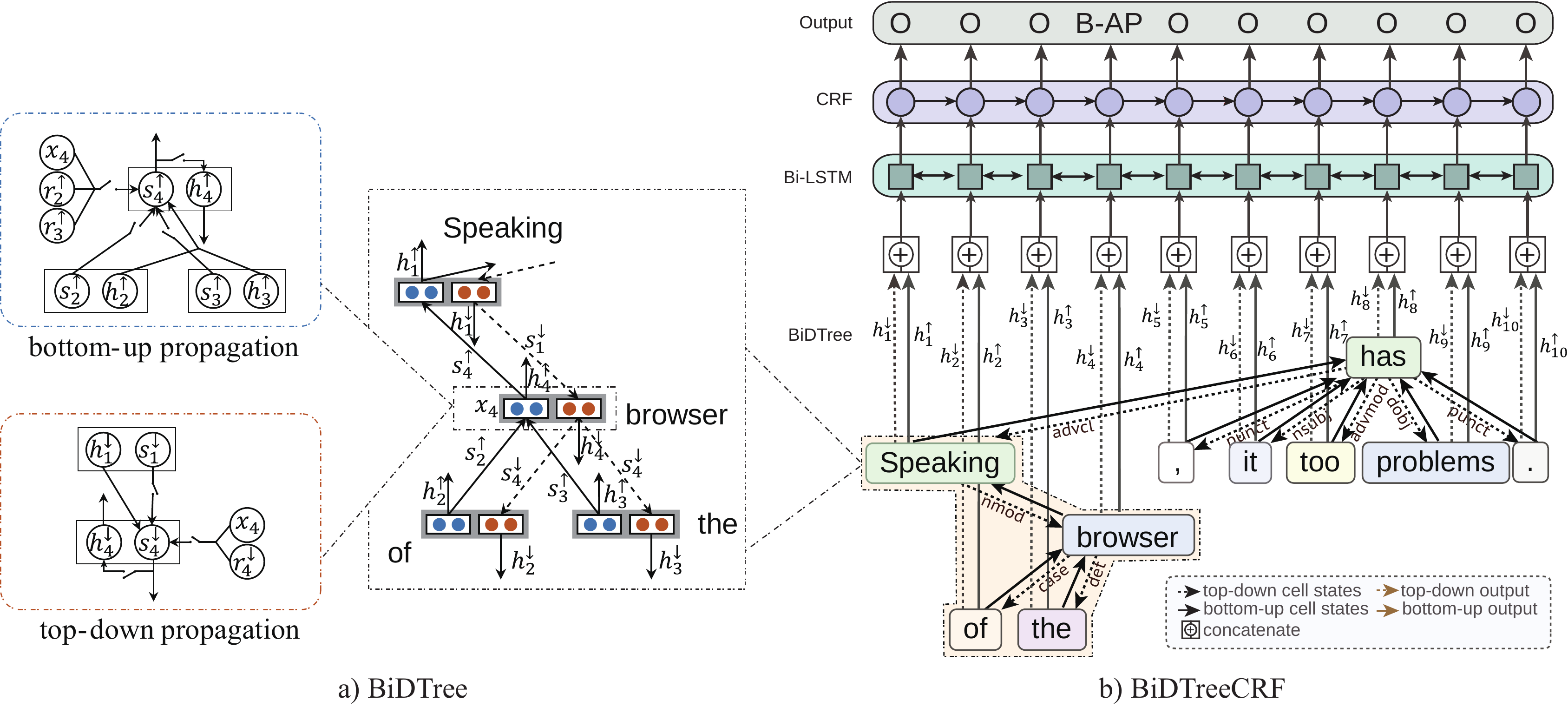}
		\caption{An illustration of the BiDTree and BiDTreeCRF architecture. Left: BiDTree architecture, including bottom-up propagation and top-down propagation; $r$ means the syntactic relation (e.g., \emph{nmod}, \emph{case}, and \emph{det}); $x$ is the word; $s$ and $h$ denote cell memory and hidden state, respectively. Right: BiDTreeCRF has three modules: BiDTree, BiLSTM, and CRF.}\label{fig_framwork}
	\end{figure*}

	The architecture of the proposed framework is shown in Figure \ref{fig_framwork}. Its sample input is the dependency relations presented in Figure \ref{fig_dependency_tree}. As described in Section \ref{sec:introduction}, BiDTreeCRF consists of three modules (or components): BiDTree, BiLSTM, and CRF. These modules will be described in details in Sections \ref{sec:bideptree} and \ref{sec:bilstmcrf}.
	
	\subsection{Problem Statement}
	\label{sec:problem}
	We are given a review sentence from a particular domain, denoted by $S = \{w_1, w_2, \dots, w_i, \dots, w_N\}$, where $N$ is the sentence length. For any word $w_i \in S$, the task of ATE is to find a label $t_i \in \mathrm{T}$ corresponding to it, where $\mathrm{T}=\{$B-AP, I-AP, O$\}$. ``B-AP", ``I-AP", and ``O" stand for the beginning of an aspect term, inside of an aspect term, and other words, respectively. For example, ``\emph{The}/O \emph{picture}/B-AP \emph{quality}/I-AP \emph{is}/O \emph{very}/O \emph{good}/O \emph{.}/O" is a sentence with labels (or tags), where the aspect term is \emph{picture quality}. This BIO encoding scheme is widely used in NLP tasks and such tasks are often solved using CRF based methods \cite{Liu2015FineGrained,Wang2016a,Irsoy2013,Irsoy2014}.
	
	\subsection{Bidirectional Dependency Tree Network}
	\label{sec:bideptree}
	Since BiDTree is built on the dependency tree, a sentence should be converted to a dependency-based parse tree first. As the left part of Figure \ref{fig_dependency_tree} shows, each node in the dependency tree represents a word and connects to at least one other node/word. Each node has one and only one head word, e.g., \emph{Speaking} is the head of \emph{browser}, \emph{has} is the head of \emph{Speaking}, and the head word of \emph{has} is ROOT\footnote{We hide it for simplicity.}. The edge between each node and its head word is a syntactic dependency relation, e.g., \emph{nmod} between \emph{browser} and \emph{Speaking} is used for nominal modifiers of nouns or clausal predicates. Syntactic relations in Figure \ref{fig_framwork} are shown as dotted black lines.
	
	After generating a dependency tree, each word $w_i$ will be initialized with a feature vector $x_{w_i} \in \mathbb{R}^{d}$, which corresponds to a column of a pre-trained word embedding $E \in \mathbb{R}^{d \times \left|V\right|}$, where $d$ is the dimension of the word vector and $\left|V\right|$ is the size of the vocabulary. As described above, each relation of a dependency tree starts from a head word and points to its dependent words. This can be formulated as follows: The governor node $p$ and its dependent nodes $c_1, c_2, \dots, c_{n_i} \dots, c_{n_p}$ are connected by $r_{pc_1}, r_{pc_2}, \dots, r_{pc_i}, \dots, r_{pc_{n_p}}$, where $n_p$ is the number of dependent nodes belonging to $p$, and $r_{pc_i} \in \mathbb{L}$, where $\mathbb{L}$ is a set of syntactic relations such as \emph{nmod, case, det, nsubj}, and so on. The syntactic relation information not only serves as features encoded in the network but also as a guide for the selection of training weights.
	
	BiDTree works in two directions using LSTM: bottom-up LSTM and top-down LSTM. Bottom-up LSTM is shown with solid black arrows and top-down LSTM is shown with dotted black arrows at the lower portion of Figure \ref{fig_framwork}. It should be noted that they are different in not only the direction but also the governor node and dependent nodes. Specifically, each node of the top-down LSTM only owns one dependent node, but the bottom-up LSTM generally owns more than one dependent node. As shown in Formula (\ref{eq_concatenate}), we concatenate the output $h_{w_i}^{\uparrow}$ of the bottom-up LSTM and the output $h_{w_i}^{\downarrow}$ of the top-down LSTM into $h_{w_i}$ as the output of BiDTree for word $w_i$,
	\begin{equation}\label{eq_concatenate}
		\begin{aligned}
			h_{w_i} = [h_{w_i}^{\uparrow}; h_{w_i}^{\downarrow}].
		\end{aligned}
	\end{equation}
	This allows BiDTree to capture the global syntactic context.
	
	Let $C(p)=\{c_1, c_2, \dots, c_{n_i} \dots, c_{n_p}\}$, which is the set of dependent nodes of node $p$ described above. Under these symbolic instructions, the bottom-up LSTM of BiDTree firstly encodes the governor word and the related syntactic relations:
	\begin{align}
		\mathcal{T}_i &= W^{\uparrow(i)}x_{w_p} + \sum_{k \in C(p)} {W_{r^{\uparrow}(k)}^{\uparrow(i)}r_{k}^{\uparrow}}, \label{eq_bideptree_it}\\
		\mathcal{T}_o &= W^{\uparrow(o)}x_{w_p} + \sum_{k \in C(p)} {W_{r^{\uparrow}(k)}^{\uparrow(o)}r_{k}^{\uparrow}}, \label{eq_bideptree_ot}\\
		\mathcal{T}_{fk} &= W^{\uparrow(f)}x_{w_p} + W_{r^{\uparrow}(k)}^{\uparrow(f)}r_{k}^{\uparrow}, \label{eq_bideptree_ft}\\
		\mathcal{T}_u &= W^{\uparrow(u)}x_{w_p} + \sum_{k \in C(p)} {W_{r^{\uparrow}(k)}^{\uparrow(u)}r_{k}^{\uparrow}}. \label{eq_bideptree_ut}
	\end{align}
	Then, the bottom-up LSTM transition equations of BiDTree are as follows:
	\begin{align}
		i_p &= \sigma\left(\mathcal{T}_i + \sum_{k \in C(p)} {U_{r^{\uparrow}(k)}^{\uparrow(i)}h_{k}^{\uparrow}} + b^{\uparrow(i)} \right), \label{eq_bideptree_i}\\
		o_p &= \sigma\left(\mathcal{T}_o + \sum_{k \in C(p)} {U_{r^{\uparrow}(k)}^{\uparrow(o)}h_{k}^{\uparrow}} + b^{\uparrow(o)} \right), \label{eq_bideptree_o}\\
		f_{pk} &= \sigma\left(\mathcal{T}_{fk} + U_{r^{\uparrow}(k)}^{\uparrow(f)}h_{k}^{\uparrow} + b^{\uparrow(f)} \right), \label{eq_bideptree_f} \\
		u_p &= \tanh\left(\mathcal{T}_u + \sum_{k \in C(p)} {U_{r^{\uparrow}(k)}^{\uparrow(u)}h_{k}^{\uparrow}} + b^{\uparrow(u)} \right), \label{eq_bideptree_u} \\
		s_p^{\uparrow} &= i_p \odot u_p + \sum_{l \in C(p)} {f_{pl} \odot s_{l}^{\uparrow}},\\
		h_p^{\uparrow} &= o_p \odot \tanh(s_p^{\uparrow}) \label{eq_bideptree_h},
	\end{align}
	where $i_p$ is the input gate, $o_p$ is the output gate, $f_{pk}$ and $f_{pl}$ are the forget gates, which are extended from the standard LSTM \cite{Hochreiter1997,Gers1999}. $s_p^{\uparrow}$ and $s_l^{\uparrow}$ are the memory cell states, $h_p^{\uparrow}$ and $h_k^{\uparrow}$ are the hidden states, $\sigma$ denotes the logistic function, $\odot$ means element-wise multiplication, $W^{\uparrow(*)}$, $W_{r^{\uparrow}(k)}^{\uparrow(*)}$, $U_{r^{\uparrow}(k)}^{\uparrow(*)}$ are weight matrices, $b^{\uparrow(*)}$ are bias vectors, and $r^{\uparrow}(k)$ is a mapping function that maps a syntactic relation type to its corresponding parameter matrix. $*\in\{i, o, f, u\}$. Specially, the syntactic relation $r_k^{\uparrow}$ is encoded into the network like word vector $x_{w_p}$ but initialized randomly. The size of $r_k^{\uparrow}$ is the same as $x_{w_p}$ in our experiments.
	
	The top-down LSTM has the same transition equations as the bottom-up LSTM, except the direction and the number of dependent nodes. Particularly, the syntactic relation type of the top-down LSTM is opposite to that of the bottom-up LSTM, and we distinguish them by adding a prefix ``\emph{I-}", e.g., setting \emph{I-nmod} to \emph{nmod}. It leads to the difference of $r^{\downarrow}(k)$ and parameter matrices. In this paper, all weights and bias vectors of BiDTree are set to size $d \times d$ and $d$-dimensions, respectively. The output $h_{w_i}$ is thus a $2d$-dimensional vector.
	
	As an instance, we give the concrete formulas of the bottom-up propagation in Figure \ref{fig_framwork} a), which are used to calculate the output of word ``browser''. On the bottom-up direction, the word ``of'' and ``the'' are related with the target word ``browser'' by the relation ``case'' and ``det'', respectively. Thus, $x_4$ is $x_{browser}$. $r_2^{\uparrow}$ and $r_3^{\uparrow}$ mean $r_{case}$ and $r_{det}$, respectively. Likewise, the subscripts 2, 3, and 4 of $s^{\uparrow}$ and $h^{\uparrow}$ are replaced with their corresponding word ``of'', ``the'', and ``browser'' to facilitate understanding. So, the output of ``browser'' on the bottom-up direction is calculated as follows:
	\begin{align}
		\begin{split}
			\mathcal{T}_i &= W^{\uparrow(i)}x_{browser} + W_{case}^{\uparrow(i)}r_{case} + W_{det}^{\uparrow(i)}r_{det}, \\
			\mathcal{T}_o &= W^{\uparrow(o)}x_{browser} + W_{case}^{\uparrow(o)}r_{case} + W_{det}^{\uparrow(o)}r_{det}, \\
			\mathcal{T}_{f(case)} &= W^{\uparrow(f)}x_{browser} + W_{case}^{\uparrow(f)}r_{case}, \\
			\mathcal{T}_{f(det)} &= W^{\uparrow(f)}x_{browser} + W_{det}^{\uparrow(f)}r_{det}, \\
			\mathcal{T}_u &= W^{\uparrow(u)}x_{browser} + W_{case}^{\uparrow(u)}r_{case} + W_{det}^{\uparrow(u)}r_{det}, \\
			i_p &= \sigma\left(\mathcal{T}_i + U_{case}^{\uparrow(i)}h_{of}^{\uparrow} + U_{det}^{\uparrow(i)}h_{the}^{\uparrow} + b^{\uparrow(i)} \right), \\
			o_p &= \sigma\left(\mathcal{T}_o + U_{case}^{\uparrow(o)}h_{of}^{\uparrow} + U_{det}^{\uparrow(o)}h_{the}^{\uparrow} + b^{\uparrow(o)} \right), \\
			f_{p(case)} &= \sigma\left(\mathcal{T}_{f(case)} + U_{case}^{\uparrow(f)}h_{of}^{\uparrow} + b^{\uparrow(f)} \right), \\
			f_{p(det)} &= \sigma\left(\mathcal{T}_{f(det)} + U_{det}^{\uparrow(f)}h_{the}^{\uparrow} + b^{\uparrow(f)} \right), \\
			u_p &= \tanh\left(\mathcal{T}_u + U_{case}^{\uparrow(u)}h_{of}^{\uparrow} + U_{det}^{\uparrow(u)}h_{the}^{\uparrow} + b^{\uparrow(u)} \right), \\
			s_{browser}^{\uparrow} &= i_p \odot u_p + f_{p(case)} \odot s_{of}^{\uparrow} + f_{p(det)} \odot s_{the}^{\uparrow},\\
			h_{browser}^{\uparrow} &= o_p \odot \tanh(s_{browser}^{\uparrow}).
		\end{split}
	\end{align}
	The top-down propagation of ``browser'' has the same formulas but with different direction. Specifically, the word ``Speaking'' is related with the target word ``browser'' by the relation ``I-nmod''. Thus, $x_4$ is $x_{browser}$ and $r_4^{\downarrow}$ refers to $r_{I\text{-}nmod}$.
	
	The formula for BiDTree is similar to the dependency layer in \cite{Miwa2016}, and the main difference is the design of parameters of the forget gate. Their work defines a parameterization of the $k$-th forget gate $f_{pk}$ of the dependent node with parameter matrices $U_{r^{\uparrow}(k)r^{\uparrow}(l)}^{\uparrow(f)}$\footnote{Same symbols are used for easy comparison}. The whole equation corresponding to Eq. (\ref{eq_bideptree_f}) is as follows:
	\begin{equation}\label{eq_differ}
		\begin{aligned}
			f_{pk} = \sigma \left( \mathcal{T}_{fk} + \sum_{l \in C(p)} U_{r^{\uparrow}(k)r^{\uparrow}(l)}^{\uparrow(f)}h_{k}^{\uparrow} + b^{\uparrow(f)} \right).
		\end{aligned}
	\end{equation}
	
	As \citeauthor{Tai2015} mentioned in \cite{Tai2015}, for a large number of dependent nodes $n_p$, using additional parameters for flexible control of information propagation from dependent to governor is impractical. Considering the proposed framework has a variable number of typed dependent nodes, we use Eq. (\ref{eq_bideptree_f}) instead of Eq. (\ref{eq_differ}) to reduce the computation cost. Another difference between their formulas and ours is that we encode the syntactic relation into our network, namely, the second term of Eqs. (\ref{eq_bideptree_it}-\ref{eq_bideptree_ut}), which is proven effective in this paper.
	\begin{figure}[t]
		\centering
		\includegraphics[width=.32\textwidth]{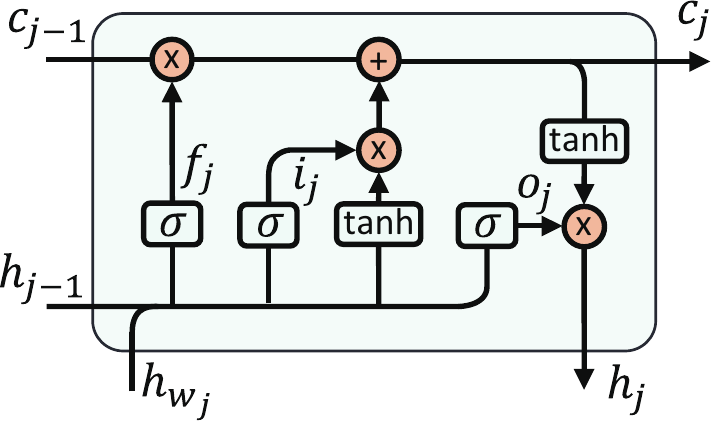}
		\caption{LSTM Unit}\label{fig_lstmcell}
	\end{figure}
	\begin{figure}[t]
		\centering
		\includegraphics[width=.28\textwidth]{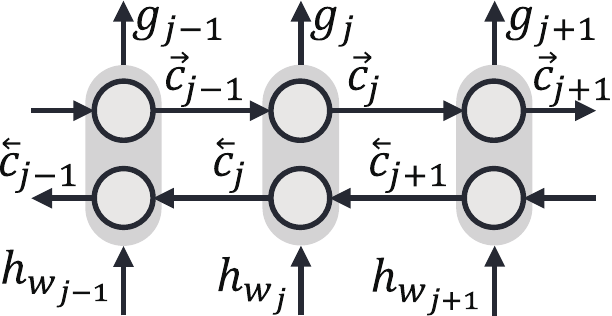}
		\caption{Bidirectional LSTM}\label{fig_bilstm}
	\end{figure}
	
	\subsection{Integration with Bidirectional LSTM}
	\label{sec:bilstmcrf}
	As the second module, BiLSTM \cite{Graves2005} keeps the sequential context of the dependency information between words. As Figure \ref{fig_lstmcell} demonstrates, the LSTM unit at $j$-th word receives the output of BiDTree $h_{w_j}$, the previous hidden state $h_{j-1}$, and the previous memory cell $c_{j-1}$ to calculate new hidden state $h_{j}$ and the new memory cell $c_{j}$ using the following equations:
	\begin{align}
		\label{eq_bilstm}
		i_j &= \sigma\left( W^{(i)}h_{w_j} + U^{(i)}h_{j-1} + b^{(i)} \right), \\
		o_j &= \sigma\left( W^{(o)}h_{w_j} + U^{(o)}h_{j-1} + b^{(o)} \right), \\
		f_j &= \sigma\left( W^{(f)}h_{w_j} + U^{(f)}h_{j-1} + b^{(f)} \right), \\
		u_j &= \tanh\left( W^{(u)}h_{w_j} + U^{(u)}h_{j-1} + b^{(u)} \right), \\
		c_j &= i_j \odot u_j + f_j \odot c_{j-1}, \\
		h_j &= o_j \odot \tanh(c_j), 
	\end{align}
	where $i_j$, $o_j$, $f_j$ are gates having the same meanings as their counterparts in BiDTree, $W^{(*)}$ with size $d \times 2d$, $U^{(*)}$ with size $d \times d$ are weight matrices, and $b^{(*)}$ are $d$-dimensional bias vectors. $*\in\{i, o, f, u\}$. We also concatenate the hidden states generated by LSTM cells in both directions belonging to the same word as the output vector, which is expressed as follows:
	\begin{align}
		\label{eq_bilstm_concatnate}
		g_{j} = \left[\overrightarrow{h_j}; \overleftarrow{h_j}\right]
	\end{align}
	The architecture of BiLSTM is shown in Figure \ref{fig_bilstm}. Also, each $g_j$ is reduced to $\left|\mathrm{T}\right|$ dimensions by a full connection layer so as to pass to the subsequent layers in our implementation.
	
	\subsection{Integration with CRF}
	The learned features actually are hybrid features containing both tree-structured and sequential information. All these features are fed into the last CRF layer to predict the label of each word. Linear-chain CRF is adopted here. Formally, let $g=\{g_1, g_2, \dots, g_j, \dots, g_N \}$ represent the output features extracted by BiDTree and BiLSTM layer. The goal of CRF is to decode the best chain of labels $y=\{t_1, t_2, \dots, t_j, \dots, t_N\}$, where $t_j$ has been described in Section \ref{sec:problem}. As a discriminant graphical model, CRF benefits from considering the correlations between labels/tags in the neighborhood, which is widely used in sequence labeling or tagging tasks \cite{Huang2015,Ma2016}. Let $\mathcal{Y}(g)$ denote all possible labels and $y'\in\mathcal{Y}(g)$. The probability of CRF $p(y|g; W,b)$ is computed as follows:
	\begin{align}
		\label{eq_crf_pro}
		p(y|g; W,b)=\frac{\prod_{j=1}^{N}{\varPsi_j{(y_{j-1},y_j,g)}}}{\sum_{y'\in\mathcal{Y}(g)}{\prod_{j=1}^{N}{\varPsi_j{(y'_{j-1},y'_j,g)}}}},
	\end{align}
	where $\varPsi_j{(y',y,g)}=\exp(W_{y',y}^{T}g+b_{y',y})$ is the potential of pair $(y',y)$. $W$ and $b$ are weight and bias, respectively.
	
	Conventionally, the training process is using maximum conditional likelihood estimation. The log-likelihood is computed as follows:
	\begin{align}
		\label{eq_crf}
		L\left( W,b \right) =\sum_j{\log\ p \left( y|g;W,b \right)}.
	\end{align}
	The last labeling results are generated with the highest conditional probability:
	\begin{align}
		y^{\ast}=\mathop{\arg\max}_{y\in \mathcal{Y}(g)} \ p(y|g; W,b).
	\end{align}
	This process is usually solved efficiently by the Viterbi algorithm.
	
	\subsection{Decoding from Labeling Results}
	Once the labeling results are generated, the last step to obtain the aspect terms of the given sentence is decoding the labeled sequence. According to the mean of elements in
	$T$, it is convenient to get the aspect terms. For example, to a sentence ``$w_1$ $w_2$ $w_3$ $w_4$'', if the labeling sequence is ``B-AP B-AP I-AP O'' then (``$w_1$'', 1, 2) and (``$w_2$ $w_3$'', 2, 4) are target aspect terms. For the above triple, the first element is the real aspect term, and the second element and the last element are the beginning (inclusive) and ending (exclusive) index in the sentence, respectively. Algorithm \ref{alg_decode} gives this process in detail.
	
	\subsection{Loss and Model Training}
	\label{sec:training}
	We equivalently use the negative of $L\left( W,b \right)$ in Eq. (\ref{eq_crf}) as the error to do minimization optimization. Thus, the loss is as follows:
	\begin{align}
		\label{eq_loss}
		\mathcal{L} = -\sum_j{\log\ p \left( y|g;W,b \right)}.
	\end{align}
	Then, the loss of the entire model is:
	\begin{align}
		\label{eq_objective}
		\mathcal{J}(\Theta) = \mathcal{L} + \frac{\lambda}{2} \lVert \Theta \rVert^2,
	\end{align}
	where $\Theta$ represents the model parameters containing all weight matrices $W$, $U$ and bias vectors $b$, and $\lambda$ is the regularization parameter.
	
	We update all parameters for BiDTreeCRF from top to bottom by propagating the errors through the CRF to the hidden layers of BiLSTM and then to BiDTree via backpropagation through time (BPTT) \cite{Goller1996}. Finally, we use Adam \cite{Kingma2015} for optimization with gradient clipping. The L2-regularization factor $\lambda$ is set as $0.001$ empirically. The mini-batch size is 20 and the initial learning rate is 0.001. We also employ dropout \cite{Srivastava2014} on the outputs of BiDTree and BiLSTM layers with the dropout rate of 0.5. All weights $W$, $U$ and bias terms $b$ are trainable parameters. Early stopping \cite{Caruana2000} is used based on performance on validation sets. Its value is 5 epochs in our experiments. At the same time, initial embeddings are fine-tuned during the training process. That means word embedding will be modified by back-propagating gradients. We implement BiDTreeCRF using the TensorFlow library \cite{Abadi2016a}, and all computations are done on an NVIDIA Tesla K80 GPU. The overall procedure of BiDTreeCRF is summarized in Algorithm \ref{alg_bidtreecrf}.
	\begin{algorithm}[tp]
		\caption{Decoding from the Labeling Sequence} 
		\label{alg_decode}
		\begin{algorithmic}[1]
			\REQUIRE A labeling sequence $\tau=\{t_1, t_2, \dots, t_i, \dots\, t_N\}$, and its corresponding sentence $S = \{w_1, w_2, \dots, w_i, \dots, w_N\}$.
			\ENSURE A list of aspect term triples
			\STATE $result \gets ()$
			\STATE $temp \gets \text{``''}$
			\STATE $start \gets 0$
			\FOR{$i=1$; $i \leq N$; $i++$}
			\IF{$t_i = \text{``O''}$ \AND $temp \neq \text{``''}$} 
			\STATE $result \gets result + (w_{start:i}, start, i)$
			\STATE $temp \gets \text{``''}$
			\STATE $start \gets 0$
			\ELSE 
			\IF{$t_i = \text{``B-AP''}$}
			\IF{$temp \neq \text{``''}$}
			\STATE $result \gets result + (w_{start:i}, start, i)$
			\ENDIF
			\STATE $temp \gets t_i$
			\STATE $start \gets i$
			\ENDIF
			\ENDIF
			\ENDFOR
			\IF{$temp \neq \text{``''}$}
			\STATE $result \gets result + (w_{start:i}, start, i)$
			\ENDIF
			\RETURN\!\!{$result$}
		\end{algorithmic}
	\end{algorithm}
	\begin{algorithm}[tp]
		\caption{BiDTreeCRF Training Algorithm} 
		\label{alg_bidtreecrf}
		\begin{algorithmic}[1]
			\REQUIRE A set of review sentences $\mathcal{S}$ from a particular domain, $S = \{w_1, w_2, \dots, w_i, \dots, w_N\}$ is one of the element in $\mathcal{S}$.
			\ENSURE Learned BiDTreeCRF model
			\STATE Construct dependency trees for each sentence $S$ using Stanford Parser Package.
			\STATE Initialize all learnable parameters $\Theta$
			\REPEAT
			\STATE Select a batch of instances $\mathcal{S}_b$ from $\mathcal{S}$
			\FOR{each sentence $S \in \mathcal{S}_b$} 
			\STATE Use BiDTree (\ref{eq_concatenate}-\ref{eq_bideptree_h}) to generate $h$
			\STATE Use BiLSTM (\ref{eq_bilstm}-\ref{eq_bilstm_concatnate}) to generate $g$
			\STATE Compute $L\left( W,b \right)$ through (\ref{eq_crf_pro}-\ref{eq_crf})
			\ENDFOR
			\STATE Use the backpropagation algorithm to update parameters $\Theta$ by minimizing the objective (\ref{eq_objective}) with the batch update mode
			\UNTIL{\emph{stopping criteria is met}}
		\end{algorithmic}
	\end{algorithm}
	
	\section{Experiments}
	\label{sec_experiments}
	In this section, we conduct experiments to evaluate the effectiveness of the proposed framework.
	
	\subsection{Datasets and Experiment Setup}
	We conduct experiments using four benchmark SemEval datasets. The detailed statistics of the datasets are summarized in Table \ref{table-datasets}.
	\begin{table}[tp]
		\caption{\label{table-datasets} Datasets from SemEval; $\#S$ means the number of sentences, $\#T$ means the number of aspect terms; L-14, R-14, R-15, and R-16 are short for Laptops 2014, Restaurants 2014, Restaurants 2015 and Restaurants 2016, respectively.}
		\begin{center}
			\begin{tabular}{|c|l|l|l|l|}
				\hline
				\textbf{Datasets} & \textbf{Train} & \textbf{Val} & \textbf{Test} & \textbf{Total} \\ \hline\hline
				L-14 $_{\#S}$ & 2,945          & 100          & 800           & 3,845          \\
				R-14 $_{\#S}$ & 2,941          & 100          & 800           & 3,841          \\
				R-15 $_{\#S}$ & 1,315          & 48           & 685           & 2,048          \\
				R-16 $_{\#S}$ & 2,000          & 48           & 676           & 2,724          \\
				\hline \hline
				L-14 $_{\#T}$ & 2,304          & 54           & 654           & 3,012          \\
				R-14 $_{\#T}$ & 3,595          & 98           & 1,134         & 4,827          \\
				R-15 $_{\#T}$ & 1,654          & 57           & 845           & 2,556          \\
				R-16 $_{\#T}$ & 2,507          & 66           & 859           & 3,432          \\ \hline
			\end{tabular}
		\end{center}
	\end{table}
	L-14 and R-14 are from SemEval 2014\footnote{\href{http://alt.qcri.org/semeval2014/task4/}{\emph{http://alt.qcri.org/semeval2014/task4/}}} \cite{Pontiki2014}, R-15 is from SemEval 2015\footnote{\href{http://alt.qcri.org/semeval2015/task12/}{\emph{http://alt.qcri.org/semeval2015/task12/}}} \cite{Pontiki2015}, and R-16 is from SemEval 2016\footnote{\href{http://alt.qcri.org/semeval2016/task5/}{\emph{http://alt.qcri.org/semeval2016/task5/}}} \cite{Pontiki2016}. L-14 contains laptop reviews, and R-14, R-15, and R-16 all contain restaurant reviews. These datasets have been officially divided into three parts: A training set, a validation set, and a test set. These divisions will be kept for a fair comparison. All these datasets contain annotated aspect terms, which will be used to generate sequence labels in the experiments. We use the Stanford Parser Package\footnote{\href{https://nlp.stanford.edu/software/lex-parser.html}{\emph{https://nlp.stanford.edu/software/lex-parser.html}}} to generate dependency trees. The evaluation metric is the F1 score, the same as the baseline methods.
	
	In order to initialize word vectors, we train word embeddings with a bag-of-words based model (CBOW) \cite{Mikolov2013} on Amazon reviews\footnote{\href{http://jmcauley.ucsd.edu/data/amazon/}{\emph{http://jmcauley.ucsd.edu/data/amazon/}}} and Yelp reviews\footnote{\href{https://www.yelp.com/academic\_dataset}{\emph{https://www.yelp.com/academic\_dataset}}}, which are in-domain corpora for laptop and restaurant, respectively. The Amazon review dataset contains 142.8M reviews, and the Yelp review dataset contains 2.2M restaurant reviews. All these datasets are trained by gensim\footnote{\href{https://radimrehurek.com/gensim/models/word2vec.html}{\emph{https://radimrehurek.com/gensim/models/word2vec.html}}} which contains the implementation of CBOW. The parameter \emph{min\_count} is 10 and \emph{iter} is 200 in our experiments. We set the dimension of word vectors to 300 based on the conclusion drawn in \cite{Wang2016a}. The experimental results about dimension settings for the proposed model also showed that 300 is a suitable choice, which provides a good trade-off between effectiveness and efficiency.
	
	\subsection{Baseline Methods and Results}
	To validate the performance of our proposed model on aspect term extraction, we compare it against a number of baselines:
	\begin{itemize}
		\item \textbf{IHS\_RD}, \textbf{DLIREC(U)}, \textbf{EliXa(U)}, and \textbf{NLANGP(U)}: The top system for L-14 in SemEval Challenge 2014 \cite{Chernyshevich2014}, the top system for R-14 in SemEval Challenge 2014 \cite{Toh2014}, the top system for R-15 in SemEval Challenge 2015 \cite{Vicente2015}, and the top system for R-16 in SemEval Challenge 2016 \cite{Toh2016}, respectively. All of these systems have the same property: They are trained on a variety of lexicon and syntactic features, which is labor-intensive compared with the end-to-end fashion of neural network. U means using additional resources without any constraint, such as lexicons or additional training data.
		\item \textbf{WDEmb}: It uses word embedding, linear context embedding and dependency path embedding to enhance CRF \cite{Yin2016}.
		\item \textbf{RNCRF-O}, \textbf{RNCRF-F}: They both extract tree-structured features using a recursive neural network as the CRF input. RNCRF-O is a model trained without opinion labels. RNCRF-F is trained not only using opinion labels but also some hand-crafted features \cite{Wang2016a}.
		\item \textbf{DTBCSNN+F}: A convolution stacked neural network built on dependency trees to capture syntactic features. Its results are produced by the inference layer \cite{Ye2017}.
		\item \textbf{MIN}: MIN is a LSTM-based deep multi-task learning framework, which jointly handles the extraction tasks of aspects and opinions via memory interactions \cite{Li2017Deep}.
		\item \textbf{CMLA}, \textbf{MTCA}: CMLA is a multilayer attention network, which exploits relations between aspect terms and opinion terms without any parsers or linguistic resources for preprocessing \cite{Wang2017}. MTCA is a multi-task attention model, which learns shared information among different tasks \cite{Wang2017a}.
		\item \textbf{LSTM+CRF}, \textbf{BiLSTM+CRF}: They are proposed by \cite{Huang2015} and produce state-of-the-art (or close to) accuracy on POS, chunking and NER data sets. We borrow them for the ATE as baselines.
		\item \textbf{BiLSTM+CNN}: BiLSTM+CNN\footnote{We use this abbreviation for the sake of typesetting.} is the Bi-directional LSTM-CNNs-CRF model from \cite{Ma2016}. Compared with BiLSTM+CRF above, BiLSTM+CNN encoded char embedding by CNN and obtained state-of-the-art performance on the task of POS tagging and named entity recognition (NER). We borrow this method for the ATE as a baseline. The window size of CNN is 3, the number of filters is 30, and the dimension of char is 100.
	\end{itemize}
	
	For our proposed model, there are three variants depending on whether the weight matrices of Eqs. (\ref{eq_bideptree_it}-\ref{eq_bideptree_u}) are shared or not \footnote{The code is publicly available at \url{https://github.com/ArrowLuo/BiDTree}}. \textbf{BiDTreeCRF\#1} shares all weight matrices, namely $W_{*}^{\uparrow(i,o,f,u)}=W^{\uparrow(i,o,f,u)}$ and $U_{*}^{\uparrow(i,o,f,u)}=U^{\uparrow(i,o,f,u)}$, which means the mapping function $r^{\uparrow}(k)$ is useless. \textbf{BiDTreeCRF\#2} shares the weight matrices of Eqs. (\ref{eq_bideptree_it}-\ref{eq_bideptree_ot}, \ref{eq_bideptree_ut}) and Eqs. (\ref{eq_bideptree_i}-\ref{eq_bideptree_o}, \ref{eq_bideptree_u}) while excluding Eqs. (\ref{eq_bideptree_ft}, \ref{eq_bideptree_f}). \textbf{BiDTreeCRF\#3} keeps Eqs. (\ref{eq_bideptree_it}-\ref{eq_bideptree_u}) and does not share any weight matrices.
	\begin{table*}[tp]
		\caption{\label{table-results-compare} Comparison on F1 scores. `-'
			indicates the results were not available in their papers \protect\footnotemark.}
		\begin{center}
			\begin{tabular}{|m{6.0cm}|m{1.5cm}<{\centering}|m{1.5cm}<{\centering}|m{1.5cm}<{\centering}|m{1.5cm}<{\centering}|}
				\hline
				\textbf{Models}    & \textbf{L-14}  & \textbf{R-14}  & \textbf{R-15}  & \textbf{R-16}  \\ \hline \hline
				IHS\_RD \cite{Chernyshevich2014}            & 74.55  & 79.62          & -              & -              \\
				DLIREC(U) \cite{Toh2014}          & 73.78          & 84.01          & -              & -              \\
				EliXa(U) \cite{Vicente2015}		   & -              & -              & 70.05          & -              \\
				NLANGP(U) \cite{Toh2016}          & -              & -              & 67.12          & 72.34          \\
				\hline \hline
				WDEmb \cite{Yin2016}              & 75.16          & 84.97          & 69.73          & -              \\
				RNCRF-O \cite{Wang2016a}            & 74.52          & 82.73          & -              & -              \\
				RNCRF+F \cite{Wang2016a}            & 78.42 & 84.93          & -              & -              \\
				DTBCSNN+F \cite{Ye2017}          & 75.66          & 83.97          & -              & -              \\
				MIN \cite{Li2017Deep}                & 77.58          & -              & -              & 73.44 \\
				CMLA \cite{Wang2017}              & 77.80          & 85.29 & 70.73		  & -              \\
				MTCA \cite{Wang2017a}              & 69.14          & -              & 71.31 & 73.26          \\
				\hline \hline
				LSTM+CRF           & 73.43          & 81.80          & 66.03          & 70.31          \\
				BiLSTM+CRF         & 76.10          & 82.38          & 65.96          & 70.11          \\
				BiLSTM+CNN         & 78.97          & 83.87          & 69.64          & 73.36          \\
				BiDTreeCRF\#1      & 80.36 & 85.08 & 69.44 & 73.74                \\
				BiDTreeCRF\#2      & 80.22 & \textbf{85.31} & 68.61 & 74.01      \\
				BiDTreeCRF\#3      & \textbf{80.57} & 84.83 & \textbf{70.83} & \textbf{74.49}  \\ \hline
			\end{tabular}
		\end{center}
	\end{table*}
	\footnotetext{We report the best results from the original papers, and keep the officially divided datasets and the evaluation program the same to make the comparison fair.}
	The different types of weight sharing mean different ways of information transmission. BiDTreeCRF\#1 shares weight matrices, which indicates the dependent words of a head word are undifferentiated and the syntactic relations, e.g., \textit{nmod} and \textit{case}, are out of consideration. BiDTreeCRF\#2 treats the forget gates differently, which indicates that each dependent word is controlled by syntactic relation to  transmitting hidden state to its next node. BiDTreeCRF\#3 further treats all gates differently. The elaborate information flow under the control of syntactic relations is proved to be efficient.
	
	The comparison results are given in Table \ref{table-results-compare}. In this table, the F1 score of the proposed model is the average of 20 runs with the same hyper-parameters that have been described in Section \ref{sec:training} and are used throughout our experiments. We report the results of L-14 initialized with the Amazon Embedding. For the other datasets, we initialize with the Yelp Embedding since they are all restaurant reviews. We will also show the embedding comparison below. 
	
	Compared to the best systems in 2014, 2015 and 2016 SemEval ABSA challenges, BiDTreeCRF\#3 achieves 6.02\%, 0.82\%, 0.78\%, and 2.15\% F1 score gains over IHS\_RD, DLIREC(U), EliXa(U) and NLANGP(U) on L-14, R-14, R-15, and R-16, respectively. Specifically, BiDTreeCRF\#3 outperforms WDEmb by 5.41\% on L-14 and 1.10\% on R-15, and outperforms RNCRF-O by 6.05\%, 2.10\% for L-14 and R-14, respectively. Even compared with RNCRF+F and DTBCSNN+F which exploit additional hand-crafted features, BiDTreeCRF\#3 on L-14 and BiDTreeCRF\#2 on R-14 without other linguistic features (e.g., POS) still achieve 2.15\%, 4.91\% and 0.38\%, 1.34\% improvements, respectively. MIN is trained via memory interactions, CMLA and MTCA are designed as a multi-task model, and all of these three methods use more labels and share information among different tasks. Comparing with them, BiDTreeCRF\#3 still gives the best score for L-14 and R-16 and a competitive score for R-15 and BiDTreeCRF\#2 achieves the state-of-the-art score for R-14, although our model is designed as a single-task model. Moreover, BiDTreeCRF\#3 outperforms LSTM+CRF and BiLSTM+CRF on all datasets by 7.14\%, 3.03\%, 4.80\%, and 4.18\%, and 4.47\%, 2.45\%, 4.87\%, and 4.38\%, respectively, and these improvements are significant ($p < 0.05$). Considering the fact that BiLSTM+CRF can be seen as BiDTreeCRF\#3 without BiDTree layer, all the results support that BiDTree can extract syntactic information effectively. 
	
	As we can see, 
	\begin{table}[tp]
		\caption{\label{table-results-assemblage} F1-scores of ablation experiments on BiDTreeCRF.}
		\begin{center}
			\begin{tabular}{|m{2.3cm}|m{0.8cm}|m{0.8cm}|m{0.8cm}|m{0.8cm}|lllll}
				\hline
				\textbf{Models} & \textbf{L-14} & \textbf{R-14} & \textbf{R-15} & \textbf{R-16} \\ \hline\hline
				BiLSTM+CRF      & 76.10         & 82.38         & 65.96         & 70.11         \\
				BiDTree+CRF     & 71.29         & 81.09         & 64.09         & 67.87         \\
				DTree-up        & 78.96         & 84.47         & 68.69         & 72.42         \\
				DTree-down      & 78.46         & 84.41         & 68.75         & 72.91         \\
				BiDTreeCRF\#3   & \textbf{80.57} & \textbf{84.83} & \textbf{70.83} & \textbf{74.49}         \\ 
				\hline
			\end{tabular}
		\end{center}
	\end{table}
	\begin{figure*}[tp]
		\centering
		\begin{minipage}[b]{0.42\textwidth}
			\centering
			\includegraphics[width=.95\textwidth]{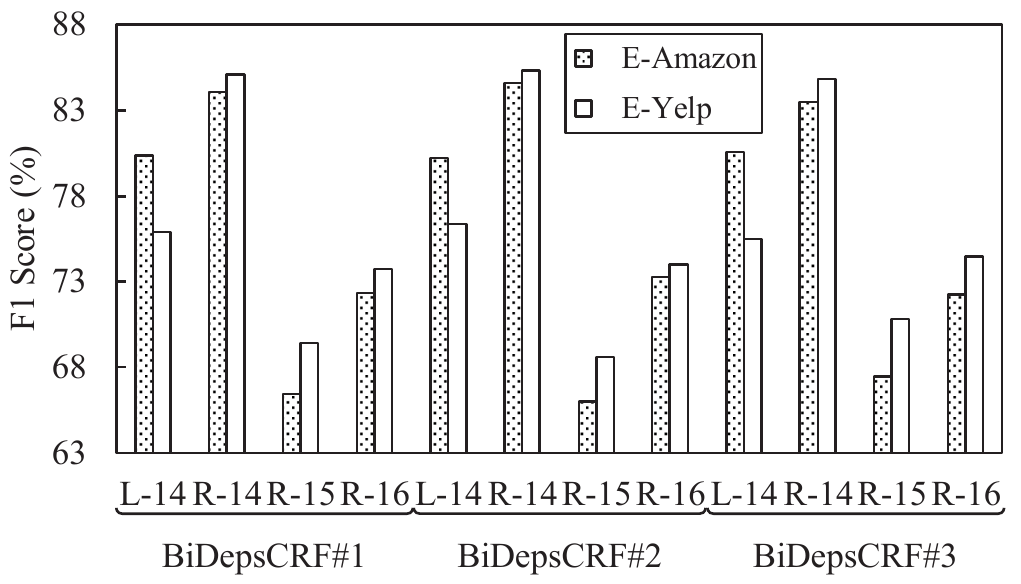}
			\caption{Amazon Embedding vs. Yelp Embedding (E-Amazon vs. E-Yelp) with syntactic relation.}
			\label{fig_CompareEmbeddingA}
		\end{minipage}
		\hspace{45.pt}
		\begin{minipage}[b]{0.42\textwidth}
			\centering
			\includegraphics[width=.95\textwidth]{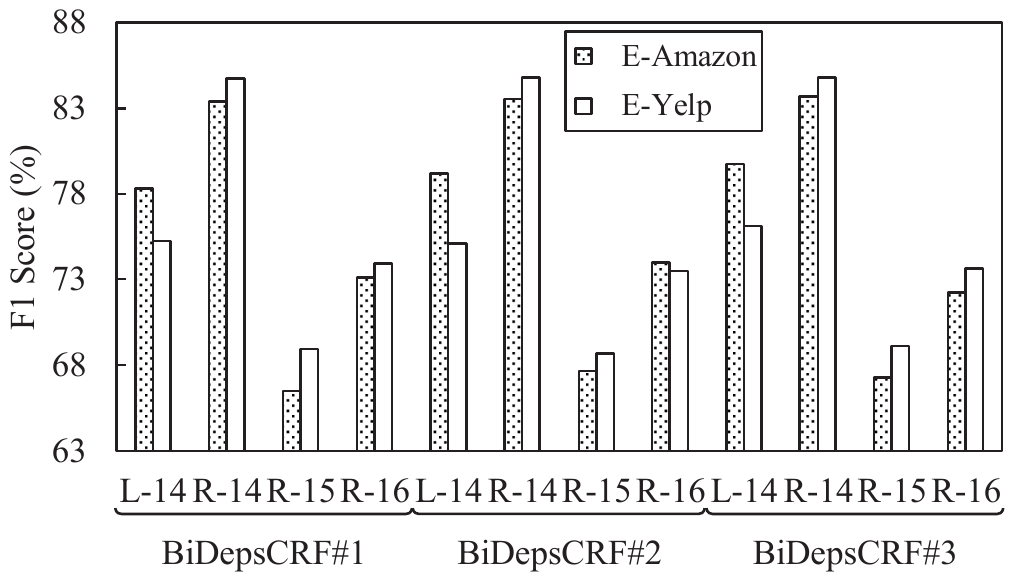}
			\caption{Amazon Embedding vs. Yelp Embedding (E-Amazon vs. E-Yelp) without syntactic relation.}
			\label{fig_CompareEmbeddingB}
		\end{minipage}
	\end{figure*}
	\begin{figure*}[tp]
		\centering
		\begin{minipage}[b]{0.42\textwidth}
			\centering
			\includegraphics[width=.95\textwidth]{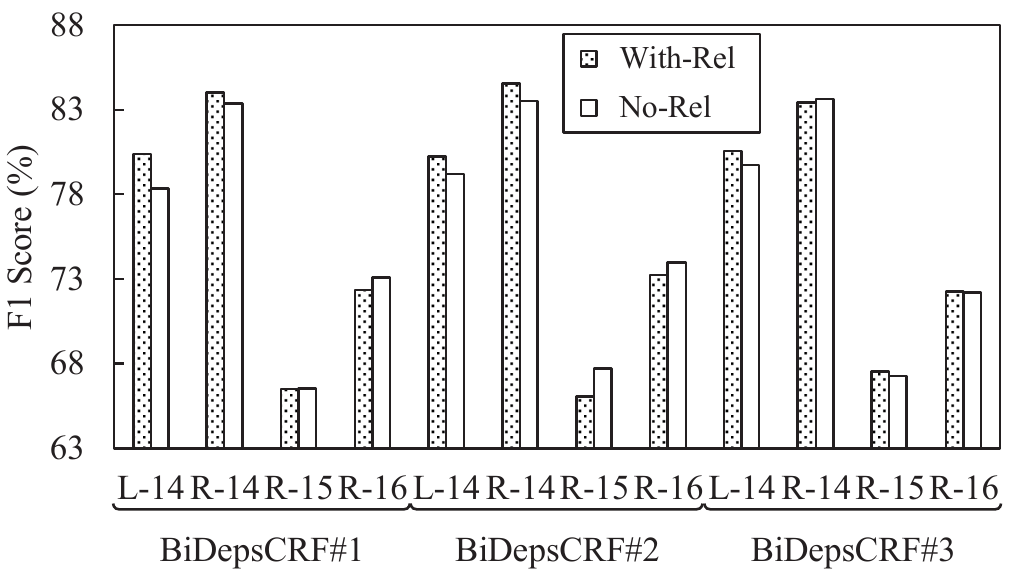}
			\caption{With syntactic relation vs. Without syntactic relation (With-Rel vs. No-Rel) with Amazon Embedding.}
			\label{fig_CompareRelA}
		\end{minipage}
		\hspace{45.pt}
		\begin{minipage}[b]{0.42\textwidth}
			\centering
			\includegraphics[width=.95\textwidth]{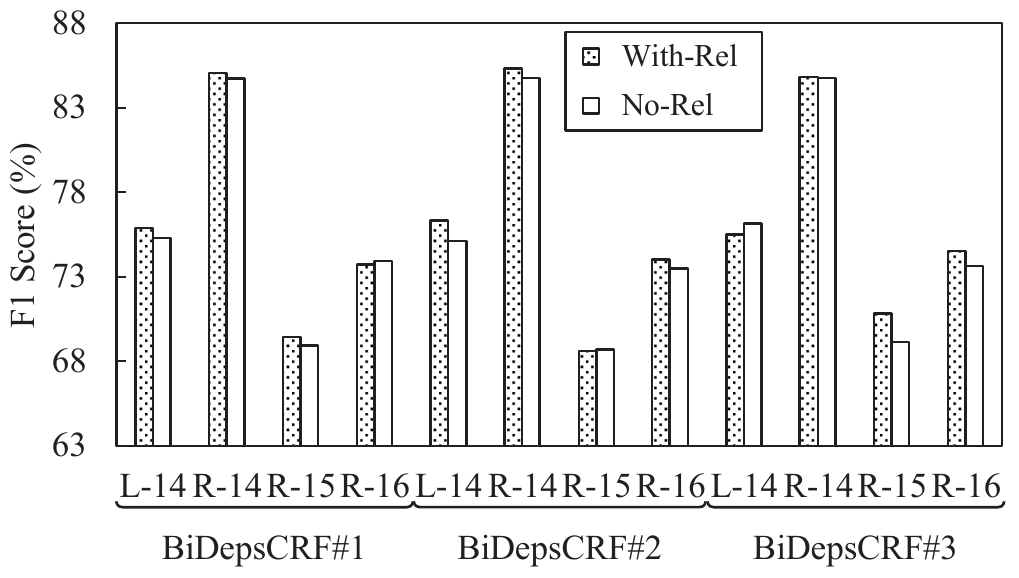}
			\caption{With syntactic relation vs. Without syntactic relation (With-Rel vs. No-Rel) with Yelp Embedding.}
			\label{fig_CompareRelB}
		\end{minipage}
	\end{figure*}
	different variants of the proposed model have different performances on the four datasets. In particular, BiDTreeCRF\#3 is more powerful than the other variants on L-14, R-15, and R-16, and BiDTreeCRF\#2 is more effective on R-15. We believe the fact that R-15 is a small dataset with some ``NULL'' aspect terms is the reason that the performance of these baselines have a small gap between them. It proves that it is a hard dataset to improve the score. Thus, it is an inspiring result though BiDTreeCRF\#3 is a little worse than MTCA without other auxiliary information (e.g., opinion terms). Besides, BiDTreeCRF\#3 outperforms BiLSTM+CNN even without char embedding. Note that we did not tune the hyperparameters of BiDTreeCRF for practical purposes because this tuning process is time-consuming.
	
	\subsection{Ablation Experiments}
	To test the effect of each component of BiDTreeCRF, the following ablation experiments on different layers of BiDTreeCRF\#3 are performed: (1) \textbf{DTree-up}: The bottom-up propagation of BiDTree is connected to BiLSTM and the CRF layer. (2) \textbf{DTree-down}: The top-down propagation of BiDTree is connected to BiLSTM and the CRF layer. (3) \textbf{BiDTree+CRF}: BiLSTM layer is not used compared to BiDTreeCRF. The initial word embeddings are the same as before. The comparison results are shown in Table \ref{table-results-assemblage}. Comparing BiDTreeCRF with DTree-up and DTree-down, it is obvious that BiDTree is more competitive than any single directional dependency network, which is the original motivation of the proposed BiDTreeCRF. The fact that BiDTreeCRF outperforms BiDTree+CRF indicates the BiLSTM layer is effective in extracting sequential information on top of BiDTree. On the other hand, the fact that BiDTreeCRF outperforms BiLSTM+CRF shows that the dependency syntactic information extracted by BiDTree is extremely useful in the aspect term extraction task. All above improvements are significant ($p < 0.05$) with the statistical t-test.
	
	\subsection{Word Embeddings \& Syntactic Relation}
	Since word embeddings are an important contributing factor for learning with less data, we also conduct comparative experiments about word embeddings. Additionally, the syntactic relation (the second terms of Eqs. (\ref{eq_bideptree_it}-\ref{eq_bideptree_ut})) is also adopted as a comparison criterion. The experimental setup, e.g., mini-batch size and learning rate, is the same as the previous setup and no other changes but word embeddings and with/without integrating syntactic relation knowledge.
	\begin{figure}[tp]
		\centering
		\includegraphics[width=.33\textwidth]{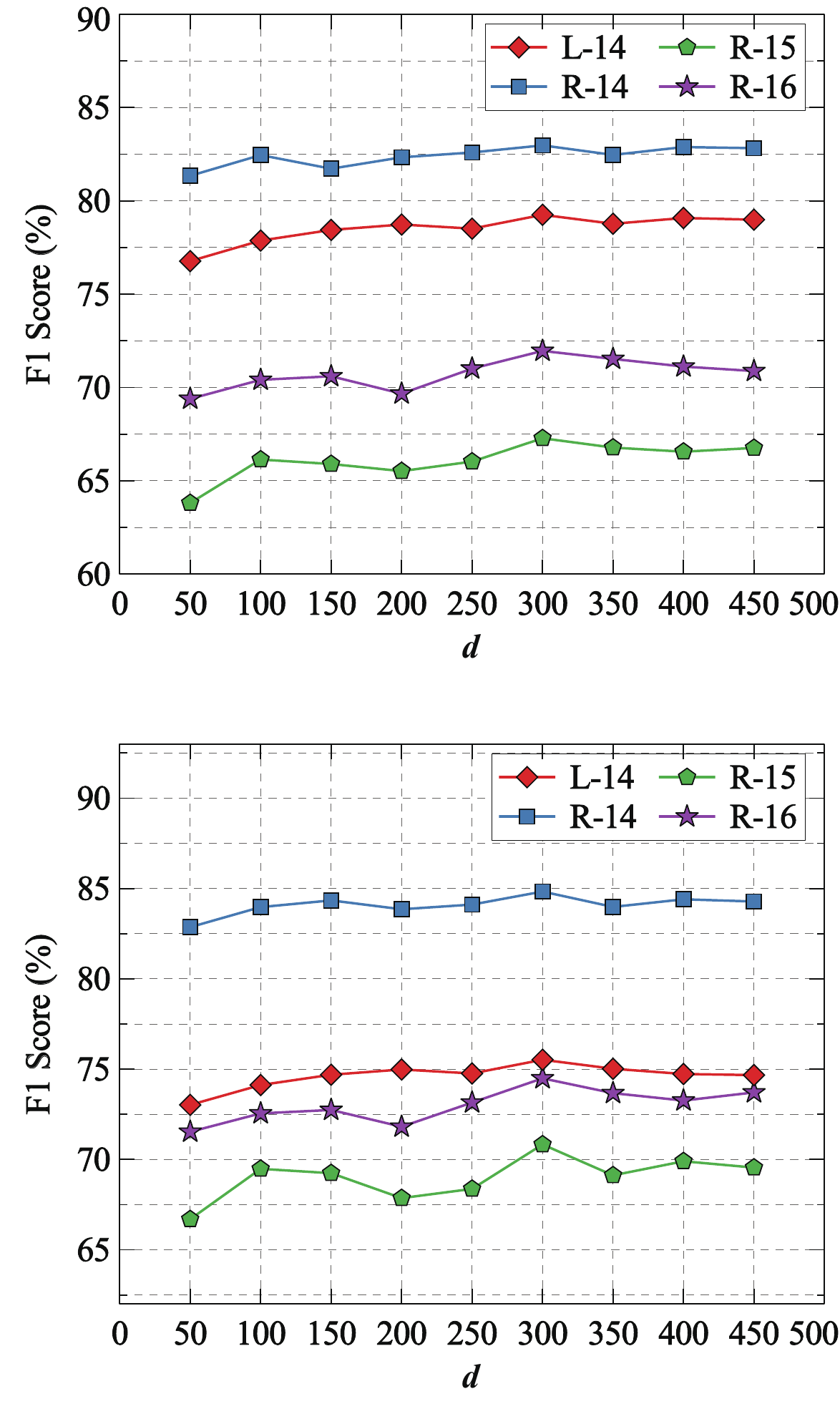}
		\caption{Sensitivity studies on word embeddings. Top: F1 Score of BiDTreeCRF\#3 with different word vector dimensions $d$ on Electronics Amazon Embedding. Bottom: F1 Score of BiDTreeCRF\#3 with different word vector dimensions $d$ on Yelp Embedding.}\label{fig_dimensionOfEmbeddings_Amazon_Yelp}
	\end{figure}
	
	Figure \ref{fig_CompareEmbeddingA} and Figure \ref{fig_CompareEmbeddingB} illustrate a comparison between Amazon Embedding and Yelp Embedding. Each figure involves three variants of BiDTreeCRF on four datasets. All of them show that Amazon Embedding is always superior to Yelp Embedding for L-14, and Yelp Embedding has an absolute advantage over Amazon Embedding for R-14, R-15, and R-16. The fact that Yelp Embedding is in-domain for restaurant and Amazon Embedding is in-domain for laptop indicates that in-domain embedding is more effective than out-domain embedding. 
	
	Figure \ref{fig_CompareRelA} and Figure \ref{fig_CompareRelB} show a comparison of different syntactic relation conditions. Figure \ref{fig_CompareRelA} is a comparison using Amazon Embedding, and Figure \ref{fig_CompareRelB} is a comparison using Yelp Embedding. The fact that the model with syntactic relation wins 7 out of 12 in Figure \ref{fig_CompareRelA} and 9 out of 12 in Figure \ref{fig_CompareRelB} comparing with the model without syntactic relation indicates the syntactic relation information is useful for performance improvement. 
	
	\subsection{Sensitivity Test}
	We conduct the sensitivity test on the dimension $d$ of word embeddings of BiDTreeCRF\#3. Different dimensions (ranging from 50 to 450, with the increment of 50) are involved. The sensitivity plots on the four datasets are given in Figure \ref{fig_dimensionOfEmbeddings_Amazon_Yelp} using Amazon Embedding and Yelp Embedding, respectively. It is worth mentioning that Amazon Embedding here is only trained from reviews of \textit{electronics} products considering the time cost. Although the score is a little lower than the embedding trained from the whole Amazon review corpus, the conclusion still holds. The figure shows that 300 is a suitable dimension size for the proposed model. It also proves the stability and robustness of our model.
	\begin{table*}[tp]
		\caption{\label{table-results-extraction} Extraction comparison between BiDTreeCRF and BiLSTM.}
		\begin{center}
			\begin{tabular}{m{5.9cm}|m{4.8cm}|m{2.5cm}|m{2.0cm}llll}
				\hline
				\textbf{Text} (The ground-truth of aspect terms is marked with bold font) & \textbf{Dependency Relationships} & \textbf{BiDTreeCRF} & \textbf{BiLSTM} \\ 
				\hline \hline
				Other than not being a fan of \textbf{click pads} (industry standard these days) and the lousy \textbf{internal speakers}, it's hard for me to find things about this notebook I don't like, especially considering the \$350 \textbf{price tag}. & \tabincell{l}{$click\xleftarrow{compound}pads$, \\ $internal\xleftarrow{amod}speakers$, \\ $price\xleftarrow{compound}tag$} & \tabincell{l}{click pads, \\ internal speakers, \\ price tag} & \tabincell{l}{internal speakers, \\ price tag} \\ 
				\hline
				\textbf{Keyboard} responds well to presses. & $Keyboard\xleftarrow{nsubj}responds$ & Keyboard & \tabincell{l}{Keyboard, \\ responds} \\ 
				\hline
				I am please with the products ease of \textbf{use}; out of the box ready; \textbf{appearance} and \textbf{functionality}. & \tabincell{l}{$ease\xrightarrow{nmod}use\xrightarrow{case}of$, \\ $appearance\xrightarrow{cc}and$, \\ $appearance\xrightarrow{conj}functionality$} & \tabincell{l}{use, \\ appearance, \\ functionality}  & \tabincell{l}{use, \\ functionality} \\ 
				\hline
				With the \textbf{softwares} supporting the use of other \textbf{OS} makes it much better. & \tabincell{l}{$use\xrightarrow{nmod}OS\xrightarrow{case}of$, \\ $the\xleftarrow{det}softwares$, \\ $softwares\xleftarrow{nsubj}supporting$} & \tabincell{l}{softwares, \\ OS} & \tabincell{l}{softwares, \\ use, \\ OS}  \\  
				\hline
				I tried several \textbf{monitors} and several \textbf{HDMI cables} and this was the case each time. & \tabincell{l}{$monitors\xrightarrow{cc}and$, \\ $monitors\xrightarrow{conj}cables$, \\ $cables\xrightarrow{compound}HDMI$ } & \tabincell{l}{monitors, \\ HDMI cables}  & HDMI cables \\
				\hline
			\end{tabular}
		\end{center}
	\end{table*}
	
	\subsection{Case Study}
	Table \ref{table-results-extraction} shows some examples from the L-14 dataset to demonstrate the effectiveness of BiDTreeCRF. The first column contains the reviews, and the corresponding aspect terms are marked with bold font. The second column describes some dependency relations related to the aspect terms. The third column and the last column are the extraction results of BiDTreeCRF and BiLSTM, respectively. On the whole, the proposed BiDTreeCRF can extract aspect terms better than BiLSTM with fewer omissions and errors. In the first example, BiLSTM misses the aspect term ``click pads'' but its inner relation is similar to the $price\xleftarrow{compound}tag$, which in the BiDTreeCRF can be considered as a significant feature. Thus BiDTreeCRF can extract it accurately. Likewise, through the relation $Keyboard\xleftarrow{nsubj}responds$, BiDTreeCRF can avoid making ``responds'' as an aspect term. For the same word ``use'' in the third example and the fourth example, one is real aspect term, and the other is not. The reason is reflected in these two relations: $ease\xrightarrow{nmod}use\xrightarrow{case}of$ and $use\xrightarrow{nmod}OS\xrightarrow{case}of$. To the final example, ``monitors'' and ``cables'' are equivalence relation because of the $monitors\xrightarrow{conj}cables$, and thus, they are extracted simultaneously by BiDTreeCRF instead of being extracted only one part of them by BiLSTM. All of the above analysis gives supporting evidence that our proposed BiDTreeCRF constructed on the dependency tree is useful and can take advantage of the relation between words to improve the ATE performance.
	
	\section{Related Work}
	\label{sec_relatedwork}
	As an important and practically very useful topic, Sentiment analysis has been extensively studied in the literature \cite{Hu2004,Cambria2016}, especially the ATE. There are several main approaches to solving the ATE problem. \citet{Hu2004} extracted aspect terms that are frequently occurring nouns and noun phrases using frequent pattern mining. \citet{Qiu2011} and \citet{Liu2015Automated} proposed to use a rule-based approach exploiting either hand-crafted or automatically generated rules about some syntactic relations between aspect terms (also called targets) and sentiment words based on the idea that opinion or sentiment must have a target \cite{Liu2012}. \citet{Chen2014Aspect} adopted the topic modeling to address the ATE, which employs some probabilistic graphical models based on Latent Dirichlet Allocation (LDA) \cite{Blei2003} and its variants. All of the above methods are based on unsupervised learning. For supervised learning, ATE is mainly regarded as a sequential labeling problem, and solved by hidden Markov models \cite{Jin2009} or CRF. However, traditional supervised methods need to design some lexical and syntactic features artificially to improve performance. Neural network is an effective approach to solve this problem.
	
	Recent work showed that neural networks can indeed achieve competitive performance on the ATE. \citet{Irsoy2013} applied deep Elman-type Recurrent Neural Network (RNN) to extract opinion expressions and showed that deep RNN outperforms CRF, semi-CRF and shallow RNN. \citet{Liu2015FineGrained} further experimented with more advanced RNN variants with fine-tune embeddings. Moreover, they pointed out that employing other linguistic features (e.g., POS) can get better results. Different from these works, \citet{poria2016aspect} used a 7-layer deep convolutional neural network (CNN) to tag each word with an aspect or non-aspect label in opinionated sentences. Some linguistic patterns were also used to improve labeling accuracy. Attention mechanism and memory interaction are also effective methods for ATE. \citet{Li2017Deep} adopted two LSTMs for jointly handling the extraction tasks of aspects and opinions via memory interactions. These LSTMs are equipped with extended memories and neural memory operations. \citet{Wang2017} proposed a multi-layer attention network to deal with aspect and opinion terms co-extraction task, which exploits the indirect relations between terms for more precise information extraction. \citet{He2017} presented an unsupervised neural attention model to discover coherent aspects. Its key idea is to exploit the distribution of word co-occurrences through the use of neural word embeddings and use an attention mechanism to de-emphasize irrelevant words during training. However, RNN and CNN based on the sequence structure of a sentence cannot effectively and directly capture the tree-based syntactic information which better reflects the syntactic properties of natural language and hence is very important to the ATE.
	
	Some tree-based neural networks have been proposed by researchers. For example, \citet{Yin2016} designed a word embedding method that considers not only the linear context but also the dependency context information. The resulting embeddings are used in CRF for extracting aspect terms. This model proves that syntactic information among words yields better performance than other representative ones for ATE. However, it involves a two-stage process, which is not an end-to-end system trained directly from the dependency path information to the final ATE tags. On the contrary, our proposed BiDTreeCRF is an end-to-end deep learning model and it does not need any hand-crafted features. \citet{Wang2016a} integrated dependency tree and CRF into a unified framework for explicit aspect and opinion terms co-extraction. However, a single directional propagation on the dependency tree is not enough to represent complete tree-structured syntactic information. Instead of the full connection on each layer of the dependency tree, we use a bidirectional propagation mechanism to extract information, which is proved to be effective in our experiments. \citet{Ye2017} proposed a tree-based convolution to capture the syntactic features of sentences, which makes it hard to keep sequential information. We fused the tree-structured and sequential information rather than only using a single representation to address the ATE efficiently.
	
	This paper is also related to several other models which are constructed on constituency trees and used to accomplish some other NLP tasks, e.g., translation \cite{Chen2017}, relation extraction \cite{Miwa2016}, relation classification \cite{Liu2015} and syntactic language modeling \cite{Tai2015,Teng2016,Zhang2016}. However, we have different models and also different applications.
	
	\section{Conclusion}
	\label{conclusion}
	In this paper, an end-to-end framework BiDTreeCRF was introduced. The framework can efficiently extract dependency syntactic information through bottom-up and top-down propagation in dependency trees. By combining the dependency syntactic information with the advantages of BiLSTM and CRF, we achieve state-of-the-art performance on four benchmark datasets without using any other linguistic features. Three variants of the proposed model have been evaluated and shown to be more effective than the existing state-of-the-art baseline methods. The distinction of these variants depends on whether they share weights during training. Our results suggest that the dependency syntactic information may also be used in aspect term and aspect opinion co-extraction, and other sequence labeling tasks. Additional linguistic features (e.g., POS) and char embeddings can further boost the performance of the proposed model.
	
	\bibliography{AspectTermExtractionRef}

\begin{thebibliography}{65}
\expandafter\ifx\csname natexlab\endcsname\relax\def\natexlab#1{#1}\fi

\bibitem[{Abadi et~al.(2016)Abadi, Agarwal, Barham et~al.}]{Abadi2016a}
Mart\'{\i}n Abadi, Ashish Agarwal, Paul Barham, et~al. 2016.
\newblock Tensorflow: Large-scale machine learning on heterogeneous distributed
  systems.
\newblock \emph{arXiv preprint arXiv:1603.04467}.

\bibitem[{Balage~Filho and Pardo(2014)}]{balage2014nilc_usp}
Pedro~Paulo Balage~Filho and Thiago Alexandre~Salgueiro Pardo. 2014.
\newblock {NIL\_CUSP}: Aspect extraction using semantic labels.
\newblock In \emph{SemEval@COLING}, pages 433--436.

\bibitem[{Blei et~al.(2003)Blei, Ng, and Jordan}]{Blei2003}
David~M Blei, Andrew~Y Ng, and Michael~I Jordan. 2003.
\newblock Latent dirichlet allocation.
\newblock \emph{JMLR}, 3(1):993--1022.

\bibitem[{Brody and Elhadad(2010)}]{Brody2010}
Samuel Brody and Noemie Elhadad. 2010.
\newblock An unsupervised aspect-sentiment model for online reviews.
\newblock In \emph{NAACL-HLT}, pages 804--812. Association for Computational
  Linguistics.

\bibitem[{Cambria(2016)}]{Cambria2016}
Erik Cambria. 2016.
\newblock \href {https://doi.org/10.1109/MIS.2016.31} {Affective computing and
  sentiment analysis}.
\newblock \emph{{IEEE} Intelligent Systems}, 31(2):102--107.

\bibitem[{Caruana et~al.(2000)Caruana, Lawrence, and Giles}]{Caruana2000}
Rich Caruana, Steve Lawrence, and C.~Lee Giles. 2000.
\newblock Overfitting in neural nets: Backpropagation, conjugate gradient, and
  early stopping.
\newblock In \emph{Advances in Neural Information Processing Systems}, pages
  402--408.

\bibitem[{Chen et~al.(2017)Chen, Huang, Chiang, and Chen}]{Chen2017}
Huadong Chen, Shujian Huang, David Chiang, and Jiajun Chen. 2017.
\newblock \href {https://doi.org/10.18653/v1/P17-1177} {Improved neural machine
  translation with a syntax-aware encoder and decoder}.
\newblock In \emph{ACL}, pages 1936--1945.

\bibitem[{Chen and Liu(2014)}]{Chen2014Topic}
Zhiyuan Chen and Bing Liu. 2014.
\newblock Topic modeling using topics from many domains, lifelong learning and
  big data.
\newblock In \emph{ICML}, pages 703--711.

\bibitem[{Chen et~al.(2014)Chen, Mukherjee, and Liu}]{Chen2014Aspect}
Zhiyuan Chen, Arjun Mukherjee, and Bing Liu. 2014.
\newblock Aspect extraction with automated prior knowledge learning.
\newblock In \emph{ACL}, pages 347--358.

\bibitem[{Chen et~al.(2013)Chen, Mukherjee, Liu, Hsu, Castellanos, and
  Ghosh}]{Chen2013}
Zhiyuan Chen, Arjun Mukherjee, Bing Liu, Meichun Hsu, Malu Castellanos, and
  Riddhiman Ghosh. 2013.
\newblock Exploiting domain knowledge in aspect extraction.
\newblock In \emph{EMNLP}, pages 1655--1667. Association for Computational
  Linguistics.

\bibitem[{Chernyshevich(2014)}]{Chernyshevich2014}
Maryna Chernyshevich. 2014.
\newblock {IHS} {R}{\&}{D} {Belarus}: Cross-domain extraction of product
  features using {CRF}.
\newblock In \emph{SemEval@COLING}, pages 309--313. Association for Computer
  Linguistics.

\bibitem[{Choi and Cardie(2010)}]{Choi2010}
Yejin Choi and Claire Cardie. 2010.
\newblock Hierarchical sequential learning for extracting opinions and their
  attributes.
\newblock In \emph{ACL}, pages 269--274. Association for Computational
  Linguistics.

\bibitem[{Gers et~al.(1999)Gers, Schmidhuber, and Cummins}]{Gers1999}
Felix~A Gers, J{\"u}rgen Schmidhuber, and Fred Cummins. 1999.
\newblock \href {https://doi.org/10.1162/089976600300015015} {Learning to
  forget: Continual prediction with lstm}.
\newblock \emph{Neural Computation}, 12(10):2451--2471.

\bibitem[{Giannakopoulos et~al.(2017)Giannakopoulos, Musat, Hossmann, and
  Baeriswyl}]{Giannakopoulos2017Unsupervised}
Athanasios Giannakopoulos, Claudiu Musat, Andreea Hossmann, and Michael
  Baeriswyl. 2017.
\newblock Unsupervised aspect term extraction with b-lstm {\&} crf using
  automatically labelled datasets.
\newblock In \emph{WASSA}, pages 180--188.

\bibitem[{Goller and Kuchler(1996)}]{Goller1996}
Christoph Goller and Andreas Kuchler. 1996.
\newblock \href {https://doi.org/10.1109/ICNN.1996.548916} {Learning
  task-dependent distributed representations by backpropagation through
  structure}.
\newblock In \emph{Proceedings of IEEE International Conference on Neural
  Networks}, pages 347--352.

\bibitem[{Graves and Schmidhuber(2005)}]{Graves2005}
Alex Graves and J{\"u}rgen Schmidhuber. 2005.
\newblock Framewise phoneme classification with bidirectional lstm and other
  neural network architectures.
\newblock \emph{Neural Networks}, 18(5-6):602--610.

\bibitem[{Hamdan et~al.(2015)Hamdan, Bellot, and B{\'{e}}chet}]{Hamdan2015}
Hussam Hamdan, Patrice Bellot, and Fr{\'{e}}d{\'{e}}ric B{\'{e}}chet. 2015.
\newblock Lsislif: {CRF} and logistic regression for opinion target extraction
  and sentiment polarity analysis.
\newblock In \emph{SemEval@NAACL-HLT}, pages 753--758. Association for Computer
  Linguistics.

\bibitem[{He et~al.(2017)He, Lee, Ng, and Dahlmeier}]{He2017}
Ruidan He, Wee~Sun Lee, Hwee~Tou Ng, and Daniel Dahlmeier. 2017.
\newblock An unsupervised neural attention model for aspect extraction.
\newblock In \emph{ACL}, pages 388--397.

\bibitem[{Hochreiter and Schmidhuber(1997)}]{Hochreiter1997}
Sepp Hochreiter and J{\"u}rgen Schmidhuber. 1997.
\newblock \href {https://doi.org/10.1162/neco.1997.9.8.1735} {Long short-term
  memory}.
\newblock \emph{Neural Computation}, 9(8):1735--1780.

\bibitem[{Hu and Liu(2004)}]{Hu2004}
Minqing Hu and Bing Liu. 2004.
\newblock \href {https://doi.org/10.1145/1014052.1014073} {Mining and
  summarizing customer reviews}.
\newblock In \emph{KDD}, pages 168--177. ACM.

\bibitem[{Huang et~al.(2015)Huang, Xu, and Yu}]{Huang2015}
Zhiheng Huang, Wei Xu, and Kai Yu. 2015.
\newblock Bidirectional lstm-crf models for sequence tagging.
\newblock \emph{arXiv preprint arXiv:1508.01991}.

\bibitem[{Irsoy and Cardie(2013)}]{Irsoy2013}
Ozan Irsoy and Claire Cardie. 2013.
\newblock Bidirectional recursive neural networks for token-level labeling with
  structure.
\newblock \emph{arXiv preprint arXiv:1312.0493}.

\bibitem[{Irsoy and Cardie(2014)}]{Irsoy2014}
Ozan Irsoy and Claire Cardie. 2014.
\newblock Opinion mining with deep recurrent neural networks.
\newblock In \emph{EMNLP}, pages 720--728. Association for Computer
  Linguistics.

\bibitem[{Jakob and Gurevych(2010)}]{Jakob2010}
Niklas Jakob and Iryna Gurevych. 2010.
\newblock Extracting opinion targets in a single-and cross-domain setting with
  conditional random fields.
\newblock In \emph{EMNLP}, pages 1035--1045. Association for Computational
  Linguistics.

\bibitem[{Jin et~al.(2009)Jin, Ho, and Srihari}]{Jin2009}
Wei Jin, Hung~Hay Ho, and Rohini~K Srihari. 2009.
\newblock \href {https://doi.org/10.1145/1553374.1553435} {A novel lexicalized
  hmm-based learning framework for web opinion mining}.
\newblock In \emph{ICML}, pages 465--472.

\bibitem[{Kingma et~al.(2014)Kingma, Ba, Kingma, and Ba}]{Kingma2015}
Diederik Kingma, Jimmy Ba, Diederik Kingma, and Jimmy Ba. 2014.
\newblock Adam: A method for stochastic optimization.
\newblock \emph{arXiv preprint arXiv:1412.6980}.

\bibitem[{Lafferty et~al.(2001)Lafferty, McCallum, and Pereira}]{Lafferty2001}
John~D. Lafferty, Andrew McCallum, and Fernando C.~N. Pereira. 2001.
\newblock Conditional random fields: Probabilistic models for segmenting and
  labeling sequence data.
\newblock In \emph{ICML}, pages 282--289.

\bibitem[{Li et~al.(2010)Li, Han, Huang, Zhu, Xia, Zhang, and Yu}]{Li2010a}
Fangtao Li, Chao Han, Minlie Huang, Xiaoyan Zhu, Ying-Ju Xia, Shu Zhang, and
  Hao Yu. 2010.
\newblock Structure-aware review mining and summarization.
\newblock In \emph{COLING}, pages 653--661. Association for Computational
  Linguistics.

\bibitem[{Li and Lam(2017)}]{Li2017Deep}
Xin Li and Wai Lam. 2017.
\newblock Deep multi-task learning for aspect term extraction with memory
  interaction.
\newblock In \emph{EMNLP}, pages 2876--2882. Association for Computational
  Linguistics.

\bibitem[{Lin and He(2009)}]{Lin2009}
Chenghua Lin and Yulan He. 2009.
\newblock Joint sentiment/topic model for sentiment analysis.
\newblock In \emph{CIKM}, pages 375--384. ACM.

\bibitem[{Liu(2012)}]{Liu2012}
Bing Liu. 2012.
\newblock Sentiment analysis and opinion mining.
\newblock \emph{Synthesis Lectures on Human Language Technologies},
  5(1):1--167.

\bibitem[{Liu et~al.(2015{\natexlab{a}})Liu, Joty, and
  Meng}]{Liu2015FineGrained}
Pengfei Liu, Shafiq Joty, and Helen Meng. 2015{\natexlab{a}}.
\newblock Fine-grained opinion mining with recurrent neural networks and word
  embeddings.
\newblock In \emph{EMNLP}, pages 1433--1443. Association for Computational
  Linguistics.

\bibitem[{Liu et~al.(2013)Liu, Gao, Liu, and Zhang}]{Liu2013}
Qian Liu, Zhiqiang Gao, Bing Liu, and Yuanlin Zhang. 2013.
\newblock A logic programming approach to aspect extraction in opinion mining.
\newblock In \emph{Proceedings of 2013 IEEE/WIC/ACM International Joint
  Conferences on Web Intelligence and Intelligent Agent Technologies}, pages
  276--283. IEEE.

\bibitem[{Liu et~al.(2015{\natexlab{b}})Liu, Gao, Liu, and
  Zhang}]{Liu2015Automated}
Qian Liu, Zhiqiang Gao, Bing Liu, and Yuanlin Zhang. 2015{\natexlab{b}}.
\newblock Automated rule selection for aspect extraction in opinion mining.
\newblock In \emph{Proceedings of the 24th International Conference on
  Artificial Intelligence}, pages 1291--1297. AAAI Press.

\bibitem[{Liu et~al.(2016)Liu, Liu, Zhang, Kim, and Gao}]{Liu2016a}
Qian Liu, Bing Liu, Yuanlin Zhang, Doo~Soon Kim, and Zhiqiang Gao. 2016.
\newblock Improving opinion aspect extraction using semantic similarity and
  aspect associations.
\newblock In \emph{AAAI}, pages 2986--2992.

\bibitem[{Liu et~al.(2015{\natexlab{c}})Liu, Wei, Li, Ji, Zhou, and
  Wang}]{Liu2015}
Yang Liu, Furu Wei, Sujian Li, Heng Ji, Ming Zhou, and Houfeng Wang.
  2015{\natexlab{c}}.
\newblock A dependency-based neural network for relation classification.
\newblock \emph{arXiv preprint arXiv:1507.04646}.

\bibitem[{Ma and Hovy(2016)}]{Ma2016}
Xuezhe Ma and Eduard~H. Hovy. 2016.
\newblock End-to-end sequence labeling via bi-directional lstm-cnns-crf.
\newblock In \emph{ACL}, pages 1064--1074. Association for Computer
  Linguistics.

\bibitem[{Mikolov et~al.(2013)Mikolov, Chen, Corrado, and Dean}]{Mikolov2013}
Tomas Mikolov, Kai Chen, Greg Corrado, and Jeffrey Dean. 2013.
\newblock Efficient estimation of word representations in vector space.
\newblock \emph{arXiv preprint arXiv:1301.3781}.

\bibitem[{Mitchell et~al.(2013)Mitchell, Aguilar, Wilson, and
  Van~Durme}]{Mitchell2013a}
Margaret Mitchell, Jacqui Aguilar, Theresa Wilson, and Benjamin Van~Durme.
  2013.
\newblock Open domain targeted sentiment.
\newblock In \emph{EMNLP}, pages 1643--1654.

\bibitem[{Miwa and Bansal(2016)}]{Miwa2016}
Makoto Miwa and Mohit Bansal. 2016.
\newblock End-to-end relation extraction using lstms on sequences and tree
  structures.
\newblock In \emph{ACL}, pages 1105--1116. Association for Computer
  Linguistics.

\bibitem[{Moghaddam and Ester(2011)}]{Moghaddam2011}
Samaneh Moghaddam and Martin Ester. 2011.
\newblock \href {https://doi.org/10.1145/2009916.2010006} {{ILDA}:
  interdependent lda model for learning latent aspects and their ratings from
  online product reviews}.
\newblock In \emph{SIGIR}, pages 665--674. ACM.

\bibitem[{Pontiki et~al.(2015)Pontiki, Galanis, Papageorgiou, Manandhar, and
  Androutsopoulos}]{Pontiki2015}
Maria Pontiki, Dimitris Galanis, Haris Papageorgiou, Suresh Manandhar, and Ion
  Androutsopoulos. 2015.
\newblock Semeval-2015 task 12: Aspect based sentiment analysis.
\newblock In \emph{SemEval@NAACL-HLT}, pages 486--495. Association for Computer
  Linguistics.

\bibitem[{Pontiki et~al.(2016)Pontiki, Galanis, Papageorgiou
  et~al.}]{Pontiki2016}
Maria Pontiki, Dimitris Galanis, Haris Papageorgiou, et~al. 2016.
\newblock Semeval-2016 task 5: Aspect based sentiment analysis.
\newblock In \emph{SemEval@NAACL-HLT}, pages 19--30. Association for Computer
  Linguistics.

\bibitem[{Pontiki et~al.(2014)Pontiki, Galanis, Pavlopoulos, Papageorgiou,
  Androutsopoulos, and Manandhar}]{Pontiki2014}
Maria Pontiki, Dimitris Galanis, John Pavlopoulos, Harris Papageorgiou, Ion
  Androutsopoulos, and Suresh Manandhar. 2014.
\newblock Semeval-2014 task 4: Aspect based sentiment analysis.
\newblock In \emph{SemEval@COLING}, pages 27--35. Association for Computer
  Linguistics.

\bibitem[{Poria et~al.(2016)Poria, Cambria, and Gelbukh}]{poria2016aspect}
Soujanya Poria, Erik Cambria, and Alexander Gelbukh. 2016.
\newblock Aspect extraction for opinion mining with a deep convolutional neural
  network.
\newblock \emph{Knowledge-Based Systems}, 108:42--49.

\bibitem[{Qiu et~al.(2011)Qiu, Liu, Bu, and Chen}]{Qiu2011}
Guang Qiu, Bing Liu, Jiajun Bu, and Chun Chen. 2011.
\newblock \href {https://doi.org/10.1162/coli\_a\_00034} {Opinion word
  expansion and target extraction through double propagation}.
\newblock \emph{Computational Linguistics}, 37(1):9--27.

\bibitem[{Shu et~al.(2017)Shu, Xu, and Liu}]{Shu2017}
Lei Shu, Hu~Xu, and Bing Liu. 2017.
\newblock \href {https://doi.org/http://aclweb.org/anthology/P17-2023}
  {Lifelong learning crf for supervised aspect extraction}.
\newblock In \emph{ACL}, pages 148--154.

\bibitem[{Socher et~al.(2011)Socher, Lin, Manning, and Ng}]{Socher2011}
Richard Socher, Cliff~C Lin, Chris Manning, and Andrew~Y Ng. 2011.
\newblock Parsing natural scenes and natural language with recursive neural
  networks.
\newblock In \emph{ICML}, pages 129--136.

\bibitem[{Srivastava et~al.(2014)Srivastava, Hinton, Krizhevsky, Sutskever, and
  Salakhutdinov}]{Srivastava2014}
Nitish Srivastava, Geoffrey~E. Hinton, Alex Krizhevsky, Ilya Sutskever, and
  Ruslan Salakhutdinov. 2014.
\newblock Dropout: a simple way to prevent neural networks from overfitting.
\newblock \emph{JMLR}, 15(1):1929--1958.

\bibitem[{Tai et~al.(2015)Tai, Socher, and Manning}]{Tai2015}
Kai~Sheng Tai, Richard Socher, and Christopher~D. Manning. 2015.
\newblock Improved semantic representations from tree-structured long
  short-term memory networks.
\newblock In \emph{ACL-AFNLP}, pages 1556--1566. Association for Computer
  Linguistics.

\bibitem[{Teng and Zhang(2016)}]{Teng2016}
Zhiyang Teng and Yue Zhang. 2016.
\newblock Bidirectional tree-structured lstm with head lexicalization.
\newblock \emph{arXiv preprint arXiv:1611.06788}.

\bibitem[{Toh and Su(2015)}]{Toh2015}
Zhiqiang Toh and Jian Su. 2015.
\newblock {NLANGP:} supervised machine learning system for aspect category
  classification and opinion target extraction.
\newblock In \emph{SemEval@NAACL-HLT}, pages 496--501. Association for Computer
  Linguistics.

\bibitem[{Toh and Su(2016)}]{Toh2016}
Zhiqiang Toh and Jian Su. 2016.
\newblock {NLANGP} at semeval-2016 task 5: Improving aspect based sentiment
  analysis using neural network features.
\newblock In \emph{SemEval@NAACL-HLT}, pages 282--288. Association for Computer
  Linguistics.

\bibitem[{Toh and Wang(2014)}]{Toh2014}
Zhiqiang Toh and Wenting Wang. 2014.
\newblock Dlirec: Aspect term extraction and term polarity classification
  system.
\newblock In \emph{Proceedings of the 8th International Workshop on Semantic
  Evaluation (SemEval 2014)}, pages 235--240.

\bibitem[{Vicente et~al.(2015)Vicente, Saralegi, and Agerri}]{Vicente2015}
I{\~{n}}aki~San Vicente, Xabier Saralegi, and Rodrigo Agerri. 2015.
\newblock {EliXa:} {A} modular and flexible {ABSA} platform.
\newblock In \emph{SemEval@NAACL-HLT}, pages 748--752. Association for Computer
  Linguistics.

\bibitem[{Wang and Wang(2008)}]{Wang2008}
Bo~Wang and Houfeng Wang. 2008.
\newblock Bootstrapping both product features and opinion words from chinese
  customer reviews with cross-inducing.
\newblock In \emph{IJCNLP}, pages 289--295.

\bibitem[{Wang et~al.(2016{\natexlab{a}})Wang, Chen, and Liu}]{Wang2016}
Shuai Wang, Zhiyuan Chen, and Bing Liu. 2016{\natexlab{a}}.
\newblock \href {https://doi.org/10.1145/2872427.2883086} {Mining
  aspect-specific opinion using a holistic lifelong topic model}.
\newblock In \emph{WWW}, pages 167--176.

\bibitem[{Wang et~al.(2017{\natexlab{a}})Wang, Pan, and Dahlmeier}]{Wang2017a}
Wenya Wang, Sinno~Jialin Pan, and Daniel Dahlmeier. 2017{\natexlab{a}}.
\newblock Multi-task coupled attentions for category-specific aspect and
  opinion terms co-extraction.
\newblock \emph{arXiv preprint arXiv:1702.01776}.

\bibitem[{Wang et~al.(2016{\natexlab{b}})Wang, Pan, Dahlmeier, and
  Xiao}]{Wang2016a}
Wenya Wang, Sinno~Jialin Pan, Daniel Dahlmeier, and Xiaokui Xiao.
  2016{\natexlab{b}}.
\newblock Recursive neural conditional random fields for aspect-based sentiment
  analysis.
\newblock In \emph{EMNLP}, pages 616--626. Association for Computational
  Linguistics.

\bibitem[{Wang et~al.(2017{\natexlab{b}})Wang, Pan, Dahlmeier, and
  Xiao}]{Wang2017}
Wenya Wang, Sinno~Jialin Pan, Daniel Dahlmeier, and Xiaokui Xiao.
  2017{\natexlab{b}}.
\newblock Coupled multi-layer attentions for co-extraction of aspect and
  opinion terms.
\newblock In \emph{AAAI}, pages 3316--3322. {AAAI} Press.

\bibitem[{Wu et~al.(2009)Wu, Zhang, Huang, and Wu}]{Wu2009}
Yuanbin Wu, Qi~Zhang, Xuanjing Huang, and Lide Wu. 2009.
\newblock Phrase dependency parsing for opinion mining.
\newblock In \emph{EMNLP}, pages 1533--1541. Association for Computational
  Linguistics.

\bibitem[{Ye et~al.(2017)Ye, Yan, Luo, and Chao}]{Ye2017}
Hai Ye, Zichao Yan, Zhunchen Luo, and Wenhan Chao. 2017.
\newblock \href {https://doi.org/10.1007/978-3-319-57529-2\_28}
  {Dependency-tree based convolutional neural networks for aspect term
  extraction}.
\newblock In \emph{PAKDD}, pages 350--362. Springer.

\bibitem[{Yin et~al.(2016)Yin, Wei, Dong, Xu, Zhang, and Zhou}]{Yin2016}
Yichun Yin, Furu Wei, Li~Dong, Kaimeng Xu, Ming Zhang, and Ming Zhou. 2016.
\newblock Unsupervised word and dependency path embeddings for aspect term
  extraction.
\newblock In \emph{IJCNN}, pages 2979--2985. {IJCAI/AAAI} Press.

\bibitem[{Zhang et~al.(2010)Zhang, Liu, Lim, and O'Brien-Strain}]{Zhang2010}
Lei Zhang, Bing Liu, Suk~Hwan Lim, and Eamonn O'Brien-Strain. 2010.
\newblock Extracting and ranking product features in opinion documents.
\newblock In \emph{COLING}, pages 1462--1470. Association for Computational
  Linguistics.

\bibitem[{Zhang et~al.(2016)Zhang, Lu, and Lapata}]{Zhang2016}
Xingxing Zhang, Liang Lu, and Mirella Lapata. 2016.
\newblock Top-down tree long short-term memory networks.
\newblock In \emph{NAACL-HLT}, pages 310--320. Association for Computational
  Linguistics.

\end{thebibliography}
	\bibliographystyle{acl_natbib}
	
\end{document}